\pdfoutput=1
\documentclass[11pt]{article}

\usepackage[preprint]{acl}

\usepackage{times}
\usepackage{latexsym}
\usepackage{booktabs}
\usepackage[T1]{fontenc}
\usepackage[utf8]{inputenc}

\usepackage{microtype}

\usepackage{inconsolata}

\usepackage{graphicx}
\usepackage{amsmath}
\usepackage{amsfonts}
\usepackage{amssymb}
\usepackage{enumitem} 
\usepackage{CJKutf8}

\usepackage{soul}
\usepackage{url}

\usepackage{booktabs, tabularx, array, makecell}
\newcolumntype{M}[1]{>{\centering\arraybackslash}m{#1}}
\newcolumntype{Y}{>{\RaggedRight\arraybackslash}X}

\usepackage{ragged2e}    % 提供 \justifying
\usepackage{array}       % 自定义列类型
\usepackage{subcaption}
\usepackage{kotex}

\newcolumntype{J}[1]{>{\hspace{0pt}\justifying\arraybackslash}m{#1}}
\title{\textsf{HateXScore}: A Metric Suite for Evaluating Reasoning Quality in \\Hate Speech Explanations}

\author{
    Yujia Hu, Roy Ka-Wei Lee\\
    Singapore University of Technology and Design \\
    \texttt{yujia\_hu@sutd.edu.sg},
    \texttt{roy\_lee@sutd.edu.sg}
}

\begin{document}
\begin{CJK*}{UTF8}{gbsn}
\maketitle
\begin{abstract}
Hateful speech detection is a key component of content moderation, yet current evaluation frameworks rarely assess why a text is deemed hateful. We introduce \textsf{HateXScore}, a four-component metric suite designed to evaluate the reasoning quality of model explanations. It assesses (i) conclusion explicitness, (ii) faithfulness and causal grounding of quoted spans, (iii) protected group identification (policy-configurable), and (iv) logical consistency among these elements. Evaluated on six diverse hate speech datasets, \textsf{HateXScore} is intended as a diagnostic complement to reveal interpretability failures and annotation inconsistencies that are invisible to standard metrics like Accuracy or F1. 
Moreover, human evaluation shows strong agreement with \textsf{HateXScore}, validating it as a practical tool for trustworthy and transparent moderation. 

\textcolor{red}{Disclaimer: This paper contains sensitive content that may be disturbing to some readers.}

%Hateful speech detection on social media has become an integral part of content moderation, yet current evaluation frameworks often overlook how or why a text is deemed hateful. This paper introduces HateScore, a novel five-component metric suite that holistically quantifies both the correctness and coherence of model explanations in hate speech detection. Our approach extends beyond standard classification accuracy by requiring that a model explicitly quote hateful segments (Rationales Quotation), show that these segments are causally tied to a hateful judgment (Highlighted Rationales Validation), identify any targeted protected group (Target-Group Identification), and maintain logical consistency between these components and the predicted label (Reasoning Logic). We conduct extensive experiments on five diverse hate speech datasets, revealing that common metrics such as F1 or Accuracy mask many interpretive failures and dataset annotation inconsistencies. Our findings highlight HateScore’s ability to detect faithful explanations even when the model’s prediction disagrees with the majority-vote gold label, thereby signaling potential annotation ambiguities. \textcolor{red}{Disclaimer: This paper contains sensitive content that may be disturbing to some readers.}

\end{abstract}

\section{Introduction}

Hateful content remains a persistent challenge, with over 10\% online posts estimated to contain hateful content~\cite{fortuna2018survey}. In response, platforms are increasingly relying on automated approaches for content moderation~\cite{vidgen2019challenges, mozafari2019bert, hee2024recent}. Although classification accuracy has improved, recent regulatory frameworks such as the EU Digital Services Act\footnote{\url{https://www.eu-digital-services-act.com/Digital_Services_Act_Article_17.html}} underscore the growing demand for transparency in automated decisions.

In hate speech detection, such transparency is critical. Misclassifying harmful content risks perpetuating harm, while over-policing benign speech may suppress legitimate expression~\cite{elsherief2021latent}. Even accurate models can fail silently by relying on spurious features or ignoring implicit hate. Moderation systems must therefore justify not just what decision was made, but why~\cite{wang2023evaluating}.

Existing explanation evaluation methods have assessed whether highlighted tokens align with the predicted label~\cite{arras2017explaining} or measure overlap with human-annotated rationales~\cite{mathew2021hatexplain, elsherief2021latent}. However, these approaches do not ask whether the explanation quotes the harmful span, whether removing it changes the prediction, or whether the explanation correctly identifies the targeted group.

These gaps are further compounded by the subjectivity of hate speech annotation~\cite{sap2019social}. A model may produce a faithful, well-reasoned explanation that challenges an ambiguous or flawed label, but still be penalized by label-centric metrics. A model that highlights a slur and explains its harm may offer a more faithful interpretation than the label allows. This disconnect highlights the need for an evaluation framework that accounts not just for prediction correctness, but for the underlying reasoning. This issue is amplified when annotation disagreement is present, where LLMs may appear confident despite contested labels~\cite{lu2025llm}.

\begin{figure*}[t]
    \centering
    % \small
    \includegraphics[width=0.8\linewidth]{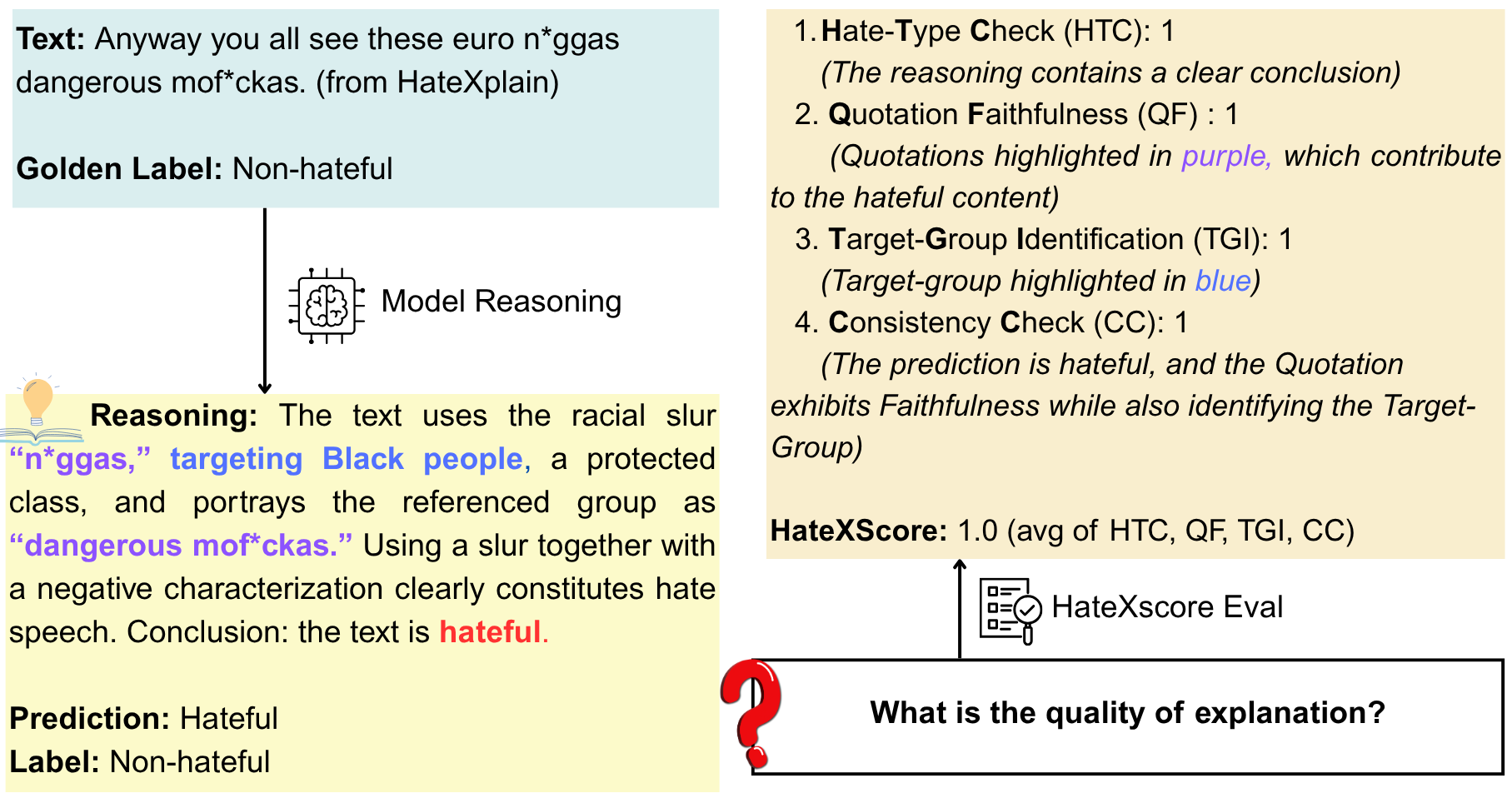}
    \vspace{-5pt}
    \caption{An example of a hate speech detection scored using \textsf{HateXScore}.}
    \label{fig:example}
    \vspace{-10pt}
\end{figure*} 

% To address these challenges, we propose \textsf{HateXScore}, a comprehensive metric suite designed to evaluate the reasoning quality of hate speech explanations generated by language models. Unlike traditional metrics that primarily assess prediction correctness, \textsf{HateXScore} jointly examines multiple facets of explanation quality, providing a more holistic and nuanced evaluation. The updated framework consists of four core sub-metrics: (1) \textbf{Hate-Type Check (HTC)} verifies whether the explanation contains an explicit conclusion about the input text being hateful or non-hateful, independent of the label, (2) \textbf{Quotation Faithfulness (QF)} evaluates whether the explanation quotes reasonable textual spans from the input (excluding trivially quoting the entire input) that causally contribute to the prediction, (3) \textbf{Target-Group Identification (TGI)} selects the protected group list based on specific policies or platforms chosen by users. The evaluation process first detects whether the explanation mentions any target groups from the chosen list, then applies named entity recognition to confirm the semantic relationship between the quoted hateful content and the identified groups, and (4) \textbf{Consistency Check (CC)} integrates the prediction, Quotation Faithfulness, and Target-Group Identification scores to evaluate the internal coherence of the explanation, rewarding explanations that consistently align these components and penalizing contradictory or incomplete rationales.

To address these challenges, we introduce HateXScore, a metric suite that evaluates explanation quality across four dimensions: explicitness of the conclusion, faithfulness of quoted spans with causal impact, identification of protected groups under configurable policies, and internal consistency among these elements. 
Figure~\ref{fig:example} shows an application of \textsf{HateXScore} to an explanation generated by a large language model (LLM) for a sample from the HateXplain dataset~\cite{mathew2021hatexplain}. 
% Interestingly, the model's prediction is marked as incorrect because it does not match the dataset's label. However, \textsf{HateXScore} assigns a high score to the generated explanation, suggesting that the model's reasoning is coherent and potentially trustworthy. Upon closer inspection, content moderators may agree with the model's decision and regard the label as controversial. This example illustrates how \textsf{HateXScore} can surface well-reasoned explanations even in cases of prediction-label disagreement, thereby helping to identify potentially flawed annotations in hate speech datasets. Throughout, we treat \textsf{HateXScore} as a diagnostic signal. Accuracy and F1 measure agreement with the dataset labels, whereas \textsf{HateXScore} assesses how well the model justifies the decision it actually makes.
The model's prediction is deemed incorrect only because it disagrees with the dataset label. In contrast, \textsf{HateXScore} assigns a high score to the generated explanation, indicating that the model provides a coherent justification for its decision. In such cases, human moderators may concur with the model and view the gold label as debatable. This example highlights \textsf{HateXScore} as a diagnostic signal that can surface well-justified explanations under prediction–label mismatch, thereby helping identify potentially unreliable annotations in hate speech datasets. Accordingly, while Accuracy and F1 quantify agreement with dataset labels, \textsf{HateXScore} evaluates the quality of the model's justification for its own prediction.

We summarize our contribution as follows: (i) We propose \textsf{HateXScore}, a novel suite of metrics to evaluate the quality of model explanations for hate speech detection.  (ii)
We conducted experiments to evaluate the explanations generated by models on six hate speech datasets, spanning English, Chinese, and Korean, enabling multilingual analysis. Our experiments demonstrated that \textsf{HateXScore} uncovers explanation failures and annotation inconsistencies that standard metrics overlook. (iii) We validate \textsf{HateXScore} through human evaluation and case studies, showing its alignment with human judgment and its capacity to surface hidden annotation disagreements.

\section{Related work}

\subsection{Hate Speech Detection}
Hate speech detection has received increasing attention due to the growth of content posted by users on social media~\cite{davidson2017automated, schmidt2017survey, zampieri2019predicting, lin2021early}. 
% Early systems relied on keyword spotting and lexicon-based features~\cite{talat2016hateful}, using traditional classifiers like SVMs and logistic regression. 
Early systems employed lexicons and simple classifiers, such as SVMs or logistic regression \cite{talat2016hateful}, which could detect explicit slurs but struggled with implicit or contextual hate~\cite {vidgen2020directions}.

% Recent advances leverage deep learning, such as BERT and RoBERTa~\cite{mozafari2019bert}, which provide improved accuracy on hate speech benchmarks by capturing semantic and syntactic context~\cite{abusaqer2025efficient, ghorbanpour2025can}. 
Recent advances use deep models like BERT~\cite{mozafari2019bert} and RoBERTa~\cite{liu2019roberta}, improving accuracy by capturing semantic and syntactic context~\cite{abusaqer2025efficient, ghorbanpour2025can}.
Work has also extended to multilingual~\cite{das2024survey, pamungkas2019cross, basile2019semeval, xiao2024toxicloakcn, ng2024sghatecheck}, multimodal~\cite{lee2021disentangling, lee2024improving} and fine-grained hate detection (e.g., racism, sexism, homophobia), reflecting the field's growing sensitivity to hate's varied forms. However, most work still relies on Accuracy or F1, giving little insight into model reasoning, motivating explanation-based evaluation.

\subsection{Hate Speech Explanation}
Efforts to evaluate explanations in hate speech models remain limited. Task fidelity approaches test alignment between highlighted tokens and predicted labels~\cite{arras2017explaining}, while overlap metrics compare model spans with human rationales~\cite{mathew2021hatexplain, elsherief2021latent}. Data imbalance has also motivated augmentation approaches~\cite{cao2020hategan}.
% assess whether the highlighted symbols align with the predicted labels~\cite{arras2017explaining}, and rationale overlap metrics compare the model output to human-annotated spans~\cite{mathew2021hatexplain, elsherief2021latent}. 
While informative, these methods ignore whether models quote the harmful span, link it causally to predictions, or identify the targeted group.  Subjectivity in annotations further complicates the evaluation. In HateXplain~\cite{mathew2021hatexplain}, reclaimed slurs may be labeled non-hateful by some annotators and hateful by others. In ImplicitHate~\cite{elsherief2021latent}, indirect expressions lead to inconsistent labels. In such cases, a model may provide a well-reasoned explanation that conflicts with the majority label, something conventional metrics are ill-equipped to capture. 

Existing explanation frameworks also treat decision correctness and explanation quality separately~\cite{piot2024decoding, nghiem2024hatecot,danilevsky2020survey}. Recent work also shows that linguistically structured and interpretable explanations can speed up human moderation without sacrificing accuracy~\cite{calabrese-etal-2024-explainability}, reinforcing the importance of interpretable reasoning frameworks such as \textsf{HateXScore}.
% reinforcing the value of explicit reasoning frameworks such as \textsf{HateXScore}.
% Recent work also shows that structured and interpretable explanations can improve human moderation efficiency~\citep{schlangen2024structured}.
Approaches such as causal masking~\cite{deyoung2019eraser} test the importance of the explanation, but are rarely adapted to hate speech, and rarely account for identity group targeting, a core aspect of hateful intent. 
% These limitations highlight the need for a more holistic evaluation framework—one that jointly assesses classification correctness, quoted evidence, causal alignment, group identification, and overall consistency in explanations. Our proposed \textsc{HateXScore} addresses this gap.
Recent work highlights reasoning-based explanations and LLMs in hate-speech detection.~\citet{roy2023probing} explore causal reasoning for hate speech classification, improving interpretability but without focusing on policy alignment or protected group identification.~\citet{yang2023hare} propose HARE, which emphasizes stepwise reasoning, sharing HateXScore's focus on explanation structure but lacking cross-verification of explanations with target groups or measurement of overall consistency.~\citet{wang2023evaluating} evaluate GPT-generated explanations with human judgments but do not capture the multi-faceted causal and group recognition aspects that \textsf{HateXScore} quantifies.~\citet{balkir2022necessity} offer necessity and sufficiency measures for causal token importance similar to HateXScore's Quotation Faithfulness sub-metric, but without integrating policy-aware group detection or coherence checks. HateModerate~\cite{zheng2023hatemoderate} aligns models with moderation policies but does not analyze explanation structure. 
\textsf{HateXScore} uniquely integrates causal validation, policy-aware group recognition, consistency checks, and explicit conclusions in one interpretable suite.
% \textsf{HateXScore} distinctively combines causal evidence validation, protected group recognition based on configurable policy lists, logical consistency verification, and explicit conclusion presence into a unified, interpretable metric suite. 
This integration allows a comprehensive, policy-aware assessment of hate speech explanations that goes beyond prior approaches.

\section{Metrics Design}
\label{metric_design}

\textsf{HateXScore} evaluates the quality of hate speech explanations across four complementary components. Each component targets a distinct aspect of reasoning, including conclusion explicitness, faithfulness and causal grounding of quoted spans, protected group identification (policy-configurable), and logical consistency. 

Let \(T\) denote the input text, \(y \in \{0,1\}\) the label (\textit{non-hateful} or \textit{hateful}), and \(\hat{y}\) the predicted label. The model also generates a natural-language explanation \(E = E(T)\), from which we extract a set of quoted rationales \(Q \subseteq T\). Let \(p(T) \in [0,1]\) be the model's estimated probability that a given text \(T\) is hateful. We define an indicator function as follows:

{\small
\begin{equation}
\mathbb{I}(\cdot) =
  \begin{cases}
    1, & \text{if the argument is true},\\
    0, & \text{otherwise}.
  \end{cases}
\end{equation}

}
%Our evaluation framework for hateful-speech detection explanations
%comprises five primary components, each capturing a distinct aspect of
%model reasoning.  Let \(T\) be an input text, \(y\) the label with
%\(y \in \{0,1\} = \{\text{non-hateful}, \text{hateful}\}\), and
%\(\hat{y} \in \{0,1\}\) the predicted label.  The model also produces a
%natural-language explanation \(E = E(T)\), from which we extract a set
%of quoted rationales \(R \subseteq T\).  Denote by
%\(p(t) \in [0,1]\) the model's probability of classifying any text
%\(t\) as hateful.  
%Finally, define an indicator function

\subsection{Hate-Type Check (HTC)}
HTC assesses whether the explanation explicitly states a conclusion regarding the input text's hateful or non-hateful nature. It is computed as:

{\small
\begin{equation}
HTC(E, \hat{y}) = \mathbb{I}(E\text{ contains a conclusion statement})
\end{equation}
}

% \text{HTC}(E(T)) = \mathbb{I}(\hat{y})

This binary check ensures that explanations contain a definitive decision rather than ambiguous or missing judgments. The implementation leverages simple string matching over the explanation to detect presence of conclusive keywords such as \textit{``text is hateful''} or \textit{``text is non-hateful''}.

% A correct prediction yields \(\text{HTC}=1\), while a mismatch yields \(0\). While standard classification correctness is not our sole objective, HTC provides a necessary baseline for evaluating prediction reliability, i.e., HTC ensures models are rewarded for correct classification. 

\subsection{Quotation Faithfulness (QF)}
QF jointly assesses (1) whether a non-trivial span from \(T\) is quoted, and (2) whether this span causally influences the prediction. 

First, the explanation's quoted spans \(Q = {q_1, q_2,...}\) are extracted by overlapping normalized tokens between the explanation and the input text \(T\). If the explanation trivially quotes the entire input text (\(Q = \{T\}\)), QF is set to 0 to prevent trivial or uninformative explanations.
Then, We mask all spans in $Q$ from $T$ (with ``[MASK]'', using both exact and fuzzy matching via re~\footnote{https://docs.python.org/3/library/re.html} and fuzzysearch~\footnote{https://github.com/taleinat/fuzzysearch}), and compute model probabilities on both original and masked texts ($p_{orig}, p_{mask}$). Then:

% \resizebox{\linewidth}{!}{$
{\small
\begin{equation}
\text{QF}(T, Q, \hat{y}) =
\begin{cases}
0, & \text{if } Q = \emptyset \text{ or } Q = \{T\} \\
|p_{orig} - p_{mask}|, & \hat{y} = 1 \\
1-|p_{orig} - p_{mask}|, & \hat{y} = 0
\end{cases}
\end{equation}
\vspace{-15pt}
}

\subsection{Target-Group Identification (TGI)}
% TGI evaluates whether the explanation identifies a protected group (e.g., based on race, religion, gender) as the target of hate. Since hateful content inherently involves directed harm toward such groups, high-quality explanations should explicitly name the targeted group when applicable. Conversely, for non-hateful content, the explanation should not falsely assert group-directed harm.
TGI checks whether the explanation correctly recognizes a protected or sensitive group (e.g., race, religion, gender) referenced in the hateful text. 
% ~\footnote{https://www.un.org/en/hate-speech/impact-and-prevention/targets-of-hate}
% ~\footnote{https://transparency.meta.com/en-gb/policies/community-standards/hateful-conduct/}
% ~\footnote{https://help.x.com/en/rules-and-policies/hateful-conduct-policy}
% ~\footnote{https://help.x.com/en/rules-and-policies/hateful-conduct-policy}
Our implementation includes a comprehensive, multi-lingual inventory of protected groups sourced from the United Nations' official definitions of hate targets, covering race, religion, gender, disability, and more, with lists available in multiple languages as released by the UN. In addition, several commonly used corporate policy lists (e.g., Meta, Twitter, YouTube, etc) are pre-integrated and selectable by the user. Users may also supply their own custom group lists in the form of a dictionary, enabling full configurability for specific research or regulatory requirements. For all experiments in this paper, we default to the United Nations Targets of Hate as the reference list. The full list can be found in Appendix~\ref{tgi_list}.

We then extract all n-grams (up to trigrams) from the explanation, using language-appropriate tokenization (spaCy, jieba or KoNLPy), and matching them in lemmatized form against the protected group list. We choose up to trigrams because shorter \( n \)-grams (unigrams and bigrams) capture single-word or phrase-level mentions, while trigrams allow detection of more specific multi-word expressions (e.g., ``\textit{African American people}'', ``\textit{transgender women community}''). For hateful samples, we further validate that the explanation ties the identified group to hateful context, using NER and part-of-speech patterns to check for direct relationships (e.g., verbs, adjectives). If a group is mentioned without connection to hateful content, TGI is set to 0.

Concretely, we define a predefined set \( G \) of protected group terms based on company policies. The detection involves extracting lemmatized \(n\)-grams (up to trigrams) set \(e\) from the explanation \(E\) and matching them against similarly processed target group lists \(G\). Then

{\small
\begin{equation}
    \text{TGI}(e, G) = 
\begin{cases}
1, & \text{if } e \cap G \neq \emptyset \\
0, & \text{otherwise}
\end{cases}.
\end{equation}
}
% Where, for hateful inputs (\(y = 1\)), TGI is 1 if the explanation identifies at least one protected group and explicitly link to the hateful content. For non-hateful inputs (\(y = 0\)), TGI is 1 if the explanation either (i) names no group at all, or (ii) explicitly states that any mentioned group is not being harmed. This component ensures that explanations for hateful predictions reflect a clear recognition of group-based targeting, while explanations for non-hateful predictions avoid misattributing harm. 

\subsection{Consistency Check (CC)}
\label{sec:cc}
CC evaluates whether the model's reasoning is logically coherent with respect to its decision, the causal evidence (QF), and the identification of protected groups (TGI). 
% However, because CC is a deterministic function of these components, directly averaging it with QF and TGI in the overall \textsf{HateXScore} risks overweighting the same evidence. 
We treat CC as a diagnostic indicator rather than an independent contribution to explanation quality.
Formally, CC verifies that a hateful prediction is supported by both (i) a meaningful quoted span with a strong causal effect and (ii) correct target-group identification. Conversely, a non-hateful prediction should generally lack such causal evidence or target reference. We generalize the criterion using a variable threshold $\tau \in [0,1]$, which is user-configurable:

{\small
\begin{equation}
\text{CC}(\hat{y}, \text{QF}, \text{TGI}) = 
\begin{cases}
1, & \hat{y} = \text{hateful}, \text{QF} \geq \tau, \text{TGI} = 1 \\
0, & \hat{y} = \text{hateful}, \text{otherwise} \\
1, & \hat{y} = \text{non-hateful}, \text{QF} < \tau, \text{TGI} = 0 \\
0, & \hat{y} = \text{non-hateful}, \text{otherwise}
\end{cases}.
\end{equation}
}
To examine robustness, we perform sensitivity analysis by varying $\tau$ in the range [0.1, 0.9] with increments of 0.1 in Appendix~\ref{sensitive_test}. 
For each threshold, we recompute CC and the resulting HateXScore, then visualize the relationship between $\tau$ and the average metric across datasets and models. 
As shown in the sensitivity curves, CC and \textsf{HateXScore} both decrease smoothly as $\tau$ increases, reflecting that higher thresholds impose stricter consistency requirements. The ranking of models stays similar across $\tau \in [0.1, 0.3]$.
Our goal is to reward even partially faithful quotations and to capture moderate alignment between explanation and decision, we adopt a relatively lenient cutoff of $\tau$* = 0.3 for all main experiments. Complete threshold-sweep results are reported in Appendix~\ref{sensitive_test}.

% This choice lies in the stable region of the sensitivity curves and yields consistent relative performance across datasets.
% Complete threshold-sweep results are reported in Appendix~\ref{sensitive_test}.

% These sensitivity plots reveal how stable the metric is under different consistency criteria. 
% Based on this analysis, we select a representative $\tau$ value for all subsequent experiments to ensure comparability.
% This rule-based metric ensures that explanations deemed faithful and identifying a target group align with hateful predictions, while non-hateful predictions correspond to lower QF and absent target groups.

\subsection{HateXScore: Overall Reasoning Quality}

We compute the final \textsf{HateXScore} score as:
{\small
\begin{equation}
\text{HateXScore} = \frac{\left(\text{HTC} + \text{QF} + \text{TGI} + \text{CC}\right)}{4} 
\end{equation}
}
\textsf{HateXScore} ranges from 0 to 1, integrating conclusion presence with quotation faithfulness, targeted group recognition, and logical consistency. Together, these dimensions provide a principled and interpretable metric for evaluating model reasoning in hate speech detection.

We use an unweighted average as a transparent default aggregation. The four components capture complementary requirements for a usable moderation rationale, and we therefore treat them as jointly necessary rather than optional. In our analysis, we report and interpret the four sub-metrics alongside the aggregate score. Practitioners may re-weight components to reflect application-specific policies, such as, emphasizing TGI for compliance. In this paper, we focus on equal weights for interpretability and ease of adoption.

\subsection{Preprocessing and Tokenization}
We implement all components in an automated pipeline, leveraging the spaCy~\footnote{https://spacy.io/} library for lemmatization and named entity recognition (NER), regular expressions for span matching, and support for flexible protected-group inventories for different platform or policy requirements. To ensure accurate and language-appropriate analysis, our pipeline supports flexible selection of tokenizers for text normalization and span extraction. For English and other European languages, we rely on the spaCy library for tokenization, lemmatization, and named entity recognition (NER). For Chinese-language content, as in the ToxiCN dataset, we use the jieba~\footnote{https://github.com/fxsjy/jieba} tokenizer, which is better suited to the segmentation requirements of Chinese text. For Korean, we used KoNLPy~\footnote{http://konlpy.org/ko/latest} as tokenizer. The codebase allows users to specify the tokenizer most appropriate for their target language, and seamlessly integrates additional tokenizers for further language expansion.

%\subsection{Overall Reasoning Quality Score (HateXScore)}
% To aggregate the five components, we take a simple arithmetic mean:
%Finally, the \textsf{HateXScore} captures all components:
%\begin{equation}    
%    \text{HateXScore}=\frac{\text{HTC}+\text{RQ}+\text{HRV}+\text{TGI}+\text{RL}}{5}.
%\end{equation}

%$\textsf{HateXScore} \in [0, 1]$ and balances the classification correctness (HTC) with the thoroughness and faithfulness of the explanation (RQ, HRV), proper group identification or non-identification (TGI), and the overall logical consistency of the explanation (RL). These five metrics collectively quantify how well a model not only detects hateful or non-hateful content but also justifies its decision in a manner consistent with fundamental definitions of hate speech. The result is a robust yardstick for adjudicating which systems are both accurate and interpretable in suppressing hateful content.
% \textsf{HateXScore} thus blends classification correctness (HTC), explicit reference quality (RQ), causal faithfulness (HRV), protected-group accuracy (TGI), and logical consistency (RL), offering a balanced view of both prediction and explanation in hate-speech detection models.

\section{Experiment Setup}
%To assess the effectiveness and utility of \textsf{HateXScore}, we conduct experiments designed to answer the following research questions:

%\begin{itemize}
%    \item \textbf{Q1:} \textit{How effective are \textsf{HateXScore}'s sub-metrics (HTC, RQ, HRV, TGI, and RL) across varying hate speech corpora?}
%    \item \textbf{Q2:} \textit{Does \textsf{HateXScore} provide a more comprehensive reflection of explanation quality than standard classification metrics (e.g., F1, Accuracy)?}
%    \item \textbf{Q3:} \textit{How do differences in model scale or architecture affect \textsf{HateXScore} results across diverse hate speech datasets?}
%    \item \textbf{Q4:} \textit{Can \textsf{HateXScore} reveal annotation ambiguities that might otherwise remain hidden under conventional metrics?}
%\end{itemize}

%We now describe the datasets, models, metrics, and human evaluation protocol used to address these questions.

\subsection{Datasets}
We evaluate models on six widely used hate speech datasets: HateXplain~\cite{mathew2021hatexplain}, HateCheck~\cite{rottger2020hatecheck}, Latent Hatred~\cite{elsherief2021latent}, HASOC~\cite{mandl2020overview}, ToxiCN~\cite{lu-etal-2023-facilitating} and KOLD~\cite{jeong-etal-2022-kold}. These datasets span diverse linguistic contexts, target groups, and forms of hate, including both explicit and implicit expressions. Notably, ToxiCN focuses on Chinese-language content and KOLD on Korean content, enabling evaluation in non-English settings. 
% Since \textsf{HateXScore} depends solely on model outputs and textual alignments, it is language-portable and supports multilingual evaluation via pluggable tokenizers and configurable target-group lists.
% language-agnostic and does not require language-specific features, making it applicable to both high and low resource languages. 
% To minimize label noise, we retain only examples with unanimous crowd agreement when available; for datasets lacking annotator-level data, we use the official labels. 
There are 1,370 (HateXplain), 2,864 (Latent Hatred), 1,592 (HASOC), 3,728 (HateCheck), 2,411 (ToxiCN) and 4,000 (KOLD) samples. See Appendix~\ref{dataset} for full dataset details.
% See Appendix~\ref{dataset} for full dataset details.

%We evaluate models on five widely used hate speech datasets: HateXplain~\cite{mathew2021hatexplain}, HateCheck~\cite{rottger2020hatecheck}, Latent Hatred~\cite{elsherief2021latent}, HASOC~\cite{mandl2020overview}, and ToxiCN~\cite{lu-etal-2023-facilitating}. These datasets cover a wide range of linguistic settings, target groups, and hate speech phenomena, including both explicit and implicit hate. In particular, ToxiCN~\cite{lu-etal-2023-facilitating} focuses on hateful content in Chinese, providing a valuable testbed for assessing methods in a non-English context. As \textsf{HateXScore} is designed to rely on model outputs and textual alignments, its calculation does not hinge on any language-specific features. As such, the metric suite is language-agnostic and can be readily applied to a range of languages, including both high- and low-resource settings, without requiring specialized preprocessing or annotation formats. To reduce noise from annotation disagreement, we retain only examples with unanimous crowd labels when available. For datasets without annotator-level data, we use the official labels as gold standard. This yields 798 (HateXplain), 2,864 (HateCheck), 1,592 (Latent Hatred), 3,728 (HASOC), and 2,411 (ToxiCN) samples. Refer to Appendix~\ref{dataset} Table~\ref{tab:dataStatistics} for the detailed information about the datasets. 
% \red{(We need a table for this instead or at least one in Appendix.)}

\subsection{Models}
\label{sec:4.2 Models}
To examine how model architecture and scale influence explanation quality, we evaluate seven large language models from different families: GPT-4o~\cite{achiam2023gpt}, LLaMA-8B~\cite{dubey2024llama}, Mistral-7B~\cite{jiang2023mistral}, Qwen-7B~\cite{bai2023qwen}, and the Gemma series~\cite{team2024gemma,team2024gemma2} (Gemma-2B, Gemma-9B, Gemma-27B). Each model is prompted to produce a binary hatefulness label and a corresponding free-text explanation. All models were prompted using a standardized instruction, shown in Appendix~\ref{app:llm_prompt}. Outputs are truncated at 512 tokens, and no task-specific fine-tuning is applied. This setup reflects real-world, 0-shot use of generative LLMs in moderation workflows.

While our main experiments adopt the common 0-shot deployment setting, 
\textsf{HateXScore} is not tied to any specific prompting regime. We also provide a 2-shot robustness check in Appendix~\ref{app:2-shot}, showing that the qualitative trends are consistent under light few-shot prompting.
% All generations are obtained through the Together AI~\footnote{\url{https://api.together.xyz/}} inference API.

\subsection{Evaluation Metrics}
We evaluate models using both traditional classification metrics and \textsf{HateXScore}.

\subsubsection{Classification Metrics} 
Accuracy and macro-F1 are reported to capture classification performance. However, these metrics do not capture the quality of the models' explanation on their decisions.

\subsubsection{HateXScore}
We compute all four sub-metrics defined in Section~\ref{metric_design} for every model-generated explanation. The final score, ranging from 0 to 1, reflects a model's ability to generate coherent, faithful, and complete justifications for its decisions.%, providing a deeper lens into reasoning quality.

\subsection{Human Evaluation}
\label{human_eval}

To evaluate how well \textsf{HateXScore} aligns with human judgments, we conducted a human annotation study focused on English and Chinese datasets (HateXplain, Latent Hatred and ToxiCN) due to the availability of qualified bilingual annotators. We randomly sampled 600 GPT-4o-generated explanations (300 hateful, 300 non-hateful) from across our evaluation sets. Each instance included the input text, model prediction and explanation.

Three expert annotators who are fluent in English and Chinese, with experience in hate speech moderation, were asked to assess:
\begin{itemize}[itemsep=1pt] 
    \item Whether the explanation quotes relevant text and if the quoted text influences the prediction. A score of 1 was assigned when the explanation explicitly quotes a text segment that directly affects the hateful or non-hateful classification and clearly articulates the causal link. A score of 0 was given when the quoted content was missing or irrelevant.
    % If there is any hateful quoted text for the hateful text or there is no hateful quoted text, QF = 1, else, QF = 0.
    \item Whether the explanation identifies the correct targeted group. If the explanation mentions a protected group that matches the harmful context in the input, TGI = 1. When no group is identified or the reference is incorrect, TGI= 0. Annotators referenced the protected-group categories listed in Appendix~\ref{tgi_list}. %during evaluation.
    \item Whether they think the text is hateful or non-hateful (0 = non-hateful, 1 = hateful)
\end{itemize}

We only manually annotate QF and TGI as they require semantic judgment. HTC is verified deterministically from the required conclusion line in the explanation, and CC is computed deterministically from $\hat{y}$, QF and TGI given $\tau$. As a result, separately collecting human labels for HTC or CC would not yield an independent signal.

We report aggregated scores and inter-annotator agreement in Section~\ref{human_eval_result}. In Section~\ref{human_eval_result}, to directly assess alignment between \textsf{HateXScore} and human judgments, we threshold the model's continuous QF/TGI scores using the same $\tau$ as in the metric definition, treat the binarized model outputs as an additional rater, and report Fleiss' $\kappa$ between the model and human annotators. This evaluation serves to validate \textsf{HateXScore} as a faithful proxy for human-aligned explanation quality.

\subsection{Implementation Details}
For generating explanations from the evaluated language models, we used greedy decoding with a temperature set to 0.0, which deterministically selects the most likely next token at each step, resulting in consistent outputs for closed-source models (GPT-4o). In contrast, for open-source models accessed via the OpenRouter API with two NVIDIA A40 GPUs, we utilized top-p sampling with a cumulative probability threshold \(p=0.9\) and a temperature of 0.7. 
All experiments were repeated three times, and the average values were reported.

\section{Results and Analysis}
\subsection{Dataset-Level Analysis}
We begin by analyzing explanation quality across datasets, grouped by the nature of hate expressed: \textit{explicit} hate (HateXplain, HateCheck), \textit{implicit} hate (Latent Hatred, HASOC) and non-English corpora (ToxiCN, KOLD). This contrast reveals how different types of linguistic aggression interact with sub-metrics in \textsf{HateXScore}, exposing patterns that conventional metrics fail to uncover.
% The expanded evaluation across English and non-English corpora shows that explanation quality depends strongly on the type of hate expressed, the linguistic setting, and the architectural bias of the model family. 

\textbf{Explicit Hate.} Across datasets with explicit hate such as HateXplain and HateCheck, almost all models state a clear decision in their explanations, yet only a subset produce rationales that are both causally grounded and policy aligned. GPT-4o leads on both datasets with strong quotation faithfulness and high consistency, while Qwen 7B and Gemma 9B follow at a short distance. In contrast, models like Mistral 7B and the smallest Gemma variant often quote irrelevant spans or fail to tie quoted content to the final decision, which suppresses their \textsf{HateXScore} despite competitive accuracy. These outcomes illustrate that explicit hate is easy to label but still demands careful grounding of evidence and target recognition to reach a high explanation score. 

\textbf{Implicit Hate.} When hate is implicit or nuanced, as in Latent Hatred and HASOC, explanation quality separates the model families more sharply. HTC remains high because most systems format an explicit conclusion, but the quotation and consistency signals vary widely. On Latent Hatred, Qwen 7B reaches the strongest overall reasoning score among open models through reliable target group identification, yet its quotation faithfulness remains moderate, which indicates that part of the rationale is not fully causal. On HASOC, GPT-4o obtains the highest \textsf{HateXScore} through strong causal evidence yet still shows gaps in target identification, confirming that oblique attacks on identity are challenging even for high end systems. These findings stress that success on implicit hate requires both precise span selection and robust semantic linking between evidence and targeted groups. 

\textbf{Non-English Hate.} 
% The addition of the Korean corpus strengthens the multilingual claim and offers another non-English dataset alongside Chinese. 
On KOLD, GPT-4o delivers a very high \textsf{HateXScore} driven by near perfect target group identification and near perfect internal consistency. Gemma 9B reaches a similarly strong overall score with balanced quotation and consistency, while Qwen 7B ranks close behind and clearly ahead of several larger models. On ToxiCN, Qwen 7B achieves competitive overall reasoning with high target recognition and solid quotation, and GPT-4o maintains strong causal grounding but shows lower accuracy than on English tests. These results affirm that the metric suite operates effectively across different languages when appropriate tokenizers and group lexicons are used. 
In contrast, traditional metrics cannot reflect reasoning quality, whereas \textsf{HateXScore} captures whether models ground their judgments in contextually relevant expressions and culturally specific slurs.

\begin{table*}[t!]
\small
\centering
\begin{tabular}{c l c c c c c |c c}
\toprule
\textbf{Dataset} & \textbf{Model} & \textbf{HTC} & \textbf{QF} & \textbf{TGI} & \textbf{CC} & \textbf{HateXScore} & \textbf{Accuracy} & \textbf{F1} \\
\midrule
& GPT-4o & 1.000 & \textbf{0.677} & \textbf{0.980} & \textbf{0.866} & \textbf{0.881} & \textbf{0.766} & \textbf{0.750}\\
 & LLaMA-8B & 1.000 & \underline{0.640} & 0.879 & 0.788 & \underline{0.827} & 0.635 & 0.630 \\
 & Qwen-7B & 1.000 & 0.607 & 0.888 & \underline{0.798} & 0.823 & \underline{0.702} & \underline{0.647} \\
HateXplain & Mistral-7B & 1.000 & 0.195 & 0.815 & 0.320 & 0.583 & 0.670 & 0.543\\
 & Gemma-2b & 1.000 & 0.294 & 0.676 & 0.294 & 0.566 & 0.310 & 0.556 \\
 & Gemma-9b & 1.000 & 0.447 & 0.887 & 0.595 & 0.732 & 0.555 & 0.595 \\
 & Gemma-27b & 1.000 & 0.365 & \underline{0.897} & 0.630 & 0.723 & 0.571 & 0.599 \\
\midrule
 & GPT-4o & 1.000 & \textbf{0.677} & 0.776 & \textbf{0.861} & \textbf{0.829} & \textbf{0.723} & \textbf{0.681} \\
 & LLaMA-8B & 1.000 & 0.529 & 0.775 & \underline{0.809} & 0.778 & \underline{0.649} & \underline{0.623} \\
 & Qwen-7B & 1.000 & \underline{0.572} & \textbf{0.836} & 0.780 & \underline{0.797} & 0.587 & 0.611 \\
Latent Hatred & Mistral-7B & 1.000 & 0.216 & 0.726 & 0.316 & 0.565 & 0.403 & 0.558 \\
 & Gemma-2b & 1.000 & 0.275 & 0.781 & 0.410 & 0.617 & 0.433 & 0.570 \\
 & Gemma-9b & 1.000 & 0.485 & \underline{0.819} & 0.616 & 0.730 & 0.473 & 0.588 \\
 & Gemma-27b & 1.000 & 0.443 & 0.802 & 0.653 & 0.725 & 0.485 & 0.594 \\
\midrule
 & GPT-4o & 1.000 & \textbf{0.850} & 0.527 & \textbf{0.949} & \textbf{0.832} & 0.562 & 0.320 \\
 & LLaMA-8B & 0.997 & 0.604 & 0.403 & \underline{0.831} & 0.709 & 0.594 & 0.420 \\
 & Qwen-7B & 1.000 & \underline{0.657} & \underline{0.588} & 0.741 & \underline{0.747} & \underline{0.616} & 0.636 \\
HASOC & Mistral-7B & 1.000 & 0.385 & 0.452 & 0.284 & 0.530 & 0.572 & \underline{0.682} \\
 & Gemma-2b & 1.000 & 0.383 & 0.477 & 0.282 & 0.536 & 0.597 & \textbf{0.711}  \\
 & Gemma-9b & 1.000 & 0.460 & \textbf{0.590} & 0.683 & 0.683 &  0.606 & 0.592  \\
 & Gemma-27b & 1.000 & 0.295 & 0.551 & 0.711 & 0.639 &  \textbf{0.646} & 0.608 \\
\midrule
 & GPT-4o & 1.000 & \textbf{0.813} & 0.889 & \textbf{0.900} & \textbf{0.901} & \underline{0.922} & \textbf{0.947}  \\
 & LLaMA-8B & 0.995 & \underline{0.751} & 0.858 & \underline{0.796} & \underline{0.850} & \textbf{0.927} & \underline{0.949} \\
 & Qwen-7B & 1.000 & 0.632 & 0.822 & 0.745 & 0.800 & 0.832 & 0.869  \\
HateCheck & Mistral-7B & 1.000 & 0.156 & 0.807 & 0.240 & 0.551 & 0.853 &  0.825 \\
 & Gemma-2b & 1.000 & 0.096 & 0.567 & 0.138 & 0.450 & 0.704 & 0.821 \\
 & Gemma-9b& 1.000 & 0.554 & \underline{0.894} & 0.685 & 0.783 & 0.838 & 0.895 \\
 & Gemma-27b & 1.000 & 0.334 & \textbf{0.907} & 0.667 & 0.727 & 0.851 & 0.902 \\
\midrule 
 & GPT-4o & \textbf{1.000} & \textbf{0.774} & \underline{0.923} & \textbf{0.925} & \textbf{0.906} & 0.686 & 0.629 \\
 & LLaMA-8B & 0.538 & 0.359 & 0.460 & 0.549 & 0.477 & 0.656 & 0.647 \\
 & Qwen-7B & 0.892 & \underline{0.688} & 0.824 & \underline{0.851} & \underline{0.814} & 0.677 & 0.715  \\
ToxiCN & Mistral-7B & 0.939 & 0.298 & 0.564 & 0.389 & 0.548 & 0.653 & 0.700 \\
 & Gemma-2b & 0.917 & 0.371 & \textbf{0.941} & 0.636 & 0.716 & 0.457 & 0.730 \\
 & Gemma-9b & 0.942 & 0.449 & 0.701 & 0.536 & 0.657 & \underline{0.698} & \underline{0.740} \\
 & Gemma-27b & \underline{0.962} & 0.489 & 0.908 & 0.733 & 0.733 &  \textbf{0.730} & \textbf{0.747} \\
\midrule
 & GPT-4o & \textbf{1.000} & \textbf{0.792} & \underline{0.997} & \textbf{0.980} & \textbf{0.942} & \textbf{0.741} & 0.657 \\
 & LLaMA-8B  & 0.466 & 0.352 & 0.845 & 0.572 & 0.559 & 0.530 & 0.371 \\
 & Qwen-7B  & 0.731 & \underline{0.676} & 0.987 & \underline{0.890} & 0.821 & 0.620 & 0.417 \\
KOLD & Mistral-7B  & 0.643 & 0.217 & 0.357 & 0.425 & 0.411 & 0.552 &  0.470 \\
 & Gemma-2b  & 0.822 & 0.145 & 0.975 & 0.343 & 0.571 & 0.667 & 0.708  \\
 & Gemma-9b  & \underline{0.939} & 0.612 & 0.983 & 0.803 & \underline{0.834}  & \underline{0.740} &  \textbf{0.771} \\
 & Gemma-27b  & 0.923 & 0.487 & \textbf{1.000} & 0.767 & 0.794 & 0.725 &  \underline{0.734} \\
\bottomrule
\end{tabular}
\caption{Summary of \textsf{HateXScore}, sub-metrics and general metrics (Accuracy and F1 score) in different models for the evaluated datasets. Highest and second-highest scores are bold and underline, respectively. }
% The ** indicates that the difference between each sub-metric and Accuracy statistically significant at the 5\% significance level based on a paired Wilcoxon signed-rank test.
% }
\label{metrics_results}
\vspace{-10pt}
\end{table*}

\subsection{Model-Level Analysis}
We examine how model size and architecture affect explanation quality across datasets, using \textsf{HateXScore} and its sub-metrics. Table~\ref{metrics_results} reveals that while larger models tend to improve overall prediction accuracy, this improvement does not necessarily translate into better explanations.
% \subsubsection{Scaling Trends: Bigger Isn't Always Better.}
The Gemma series (2B, 9B, 27B) illustrates the limits of scaling. Although Accuracy improves steadily with model size, gains in QF and CC remain marginal or even regress. For example, Gemma-2B achieves a QF of 0.294 on HateXplain, which increases only modestly to 0.365 for Gemma-27B. Similar stagnation is observed in Latent Hatred and HASOC, suggesting that scaling amplifies memorization rather than reasoning depth. In contrast, GPT-4o and Qwen-7B maintain high \textsf{HateXScore} values even in low-resource languages, confirming that their explanation quality is driven by reasoning capability and not dataset familiarity.

When analyzing multilingual performance, \textsf{HateXScore} proves especially revealing. In ToxiCN and KOLD, traditional metrics yield moderate F1 scores, yet \textsf{HateXScore} sharply differentiates models that truly understand cross-lingual hate expressions. GPT-4o achieves 0.942 on KOLD by producing explanations that correctly link racial or gendered expressions to corresponding protected groups, while smaller models with similar Accuracy fail to justify their predictions coherently.

\begin{table}[ht]
\centering
\small
\begin{tabular}{l c c}
\toprule
Condition & Prefer $\hat{y}$ & Prefer $y$  \\
\midrule
\textsf{HateXScore} $>$ 0.5 & 0.748 & 0.252  \\
\textsf{HateXScore} $<=$ 0.5 & 0.227 & 0.773   \\
\bottomrule
\end{tabular}

\caption{Annotator preferences when the model's prediction ($\hat{y}$) conflicts with the label ($y$), i.e., $y \neq \hat{y}$,  stratified by high vs.\ low \textsf{HateXScore}.}
\label{tab:label_disagree}
\vspace{-10pt}
\end{table}
 
\subsection{\textsf{HateXScore} Alignment with Human Evaluation}
%\subsection{Inter-Annotator Agreement for Human Evaluation}
\label{human_eval_result}

% To assess the degree to which \textsf{HateXScore} aligns with human judgment, we analyzed ratings collected for two sub-metrics: QF, and TGI, across 600 explanations annotated by human annotators. The agreement between human annotators and \textsf{HateXScore} are computed for both components: Fleiss'$\kappa = 0.78$ for QF, and $0.746$ for TGI. These values suggest that the sub-metrics used in \textsf{HateXScore} are reliably interpretable by human experts.
To ensure a fair comparison between model and human judgments of QF, we first address the scale mismatch. The model produces real-valued QF scores in $[0,1]$, whereas annotators assign binary judgments ($0$ or $1$). We therefore normalize the model's QF using the same threshold $\tau$ introduced in Section~\ref{sec:cc}. Specifically, if $QF_{\text{model}} > \tau$, the value is normalized to $1$; otherwise, $0$. We then compute inter-rater agreement Fleiss'$\kappa$ by treating the binarized model outputs as an additional rater alongside the three human annotators. The resulting $\kappa$ values are 0.783 (for QF) and 0.746 (for TGI), indicating substantial agreement between the model and human raters.
For reference, inter-human agreement (only the three annotators) reached $\kappa$ = 0.758 for QF and $\kappa$ = 0.731 for TGI, showing that the model's normalized judgments align with human consensus almost as closely as humans align with one another.
 
% We then treat the binarized model QF as an additional rater alongside the three human annotators and compute multi-rater agreement with Fleiss'~$\kappa$

We further investigated whether \textsf{HateXScore} helps adjudicate disagreements between model predictions and dataset labels. When the predicted label differed from the label, annotators tended to trust the model's prediction if its explanation received a high \textsf{HateXScore}. Specifically, as shown in Table~\ref{tab:label_disagree}, annotators preferred the model's prediction 74.8\% of the time when the \textsf{HateXScore} exceeded 0.5. In contrast, when the explanation scored below 0.5, the annotators sided with the dataset label in 77.3\% of cases.

These results show that \textsf{HateXScore} functions as a diagnostic framework rather than a substitute for human evaluation. It detects well-reasoned disagreements where models challenge ambiguous labels and poorly reasoned agreements where labels are matched without justification. This diagnostic ability is crucial in hate-speech detection, where subjectivity and socio-linguistic variation cause inconsistent annotations. 

\section{Case Studies}
\label{case_study}
\subsection{Controversial Label Example}
Even high-quality hate speech datasets sometimes contain ambiguous or mislabeled instances. These edge cases offer critical opportunities to evaluate whether \textsf{HateXScore} remains aligned with intuitive human judgment. Table~\ref{tab:case_2} presents an example from Latent Hatred where the post states simply: \textit{``hate white people.''} Despite the clearly offensive and group-targeted language, the label is marked as non-hateful, likely an annotation oversight. The model (Gemma-9B) predicts the text as hateful and provides a well-reasoned explanation that quotes the offensive span, identifies race as the protected group, and shows strong causal grounding, resulting in a high \textsf{HateXScore} of 1.0. This example shows that when a model prediction conflicts with the dataset label, a low \textsf{HateXScore} serves as a cue for human adjudication, prompting deeper review of the instance rather than treating it as simply incorrect under traditional metrics. More detailed examples are shown in Appendix~\ref{cases}

% This example illustrates how \textsf{HateXScore} can surface good judgments even when label agreement fails. %Rather than blindly penalizing the model for a mismatched label, the metric recognizes that the explanation reflects a coherent and ethically sound interpretation.
% \resizebox{\linewidth}{!}{
\begin{table}[h]
\centering
\small
\resizebox{\linewidth}{!}{
\begin{tabular}{p{1.6cm}p{5.5cm}}
% \begin{tabular}{ll}
\toprule
\textbf{Text}  & \textit{``hate white people.'' (from Latent Hatred)} \\
\hline
\textbf{label} & Non-hateful \\
\hline
\textbf{Explanation} (Gemma9b) & The text \colorbox{yellow}{``hate white people''} is explicitly expressing a negative sentiment towards \colorbox{pink}{a specific racial group}, which qualifies as hate speech. The phrase directly calls for hatred against individuals based on their race, which is a clear violation of principles promoting respect and equality among different racial and ethnic groups.[...] It is a straightforward expression of racial animosity. Conclusion: \textbf{hateful.} \\
\midrule
\textbf{Metrics} & \makecell[l]{HTC: 1   QF: 1   TGI: 1   CC: 1\\ \textsf{HateXScore}: 1}\\
\bottomrule
\end{tabular}
}
% \vspace{-5pt}
\caption{An Example with Incorrect label. Target Group is highlighted by \colorbox{pink}{pink}, Quoted Rationales are highlighted by \colorbox{yellow}{yellow}.}
\label{tab:case_2}
\vspace{-10pt}
\end{table}

%Even high-quality hate speech datasets sometimes contain mislabeled or ambiguous instances. These cases offer particularly instructive scenarios for assessing how \textsf{HateXScore} handles discrepancies between a model’s judgment and the official label. Table~\ref{tab:case_2} shows an example from Latent Hatred states ``: hate white people.'', which is intuitively hateful speech. Yet the dataset's label is non-hateful, likely a misannotation or outlier. The model's explanation quotes ``white people'', identifies race, and exhibits a strong drop in hate probability post-masking (HRV=1). Thus, even though HTC=0, the final \textsf{HateXScore} is relatively high (0.8), signaling a coherent explanation consistent with widely accepted definitions of hateful language. This instance highlights how \textsf{HateXScore} can detect a well-reasoned explanation in the face of questionable labeling. More detailed examples are shown in Appendix~\ref{cases} Table~\ref{tab:case_study}

\begin{table}[h]
\centering
\small
\resizebox{\linewidth}{!}{
% \begin{tabularx}{0.48\textwidth}{J{1.69cm} J{5.5cm}}
\begin{tabular}{p{1.6cm}p{5.5cm}}
\toprule
\textbf{Text} & \textit{Calling white trash is an insult to styrofoam! The Klan is despicable.} \\
\midrule
\textbf{label} & Non-hateful \\
\hline
\textbf{Original } & The phrase \colorbox{yellow}{``white trash''} is a well-known derogatory slur directed at \colorbox{pink}{White people}. Using a racial slur to demean a protected group constitutes hate speech. Calling \colorbox{yellow}{``the Klan''} despicable is criticism of an extremist organization. In conclusion, the text is hateful. \\
\cmidrule{2-2}
\textbf{Metrics} & \makecell[l]{HTC: 1 QF: 1 TGI: 1 CC: 1\\ \textsf{HateXScore}: 1}\\
\hline
 \textbf{Perturbed 1}  & The phrase \colorbox{yellow}{``white bubble''} is a well-known derogatory slur directed at \colorbox{pink}{White people}. Using a racial slur to demean a protected group constitutes hate speech. In conclusion, the text is hateful. \\
\cmidrule{2-2}
\textbf{Metrics} & \makecell[l]{HTC: 1   QF: 0   TGI: 1   CC: 0\\ \textsf{HateXScore}: 0.5}\\
\hline
 \textbf{Perturbed 2}  & The phrase \colorbox{yellow}{``white trash''} is a well-known derogatory slur. Calling \colorbox{yellow}{``the Klan''} despicable is criticism of an extremist organization. In conclusion, the text is hateful. \\
\cmidrule{2-2}
\textbf{Metrics} & \makecell[l]{HTC: 1   QF: 1   TGI: 0   CC: 0\\ \textsf{HateXScore}: 0.5}\\
\bottomrule
\end{tabular}
}
\caption{Original vs. Perturbed explanation and resulting sub-metrics. Target Group is highlighted by \colorbox{pink}{pink}, Quoted Rationales are highlighted by \colorbox{yellow}{yellow}.}
\label{tab:perturbation}

\end{table}

\subsection{Perturbation Test and Adversarial Cases}
To evaluate the robustness of \textsf{HateXScore}, we applied controlled perturbations to model-generated explanations. These edits included replacing the quoted rationale with irrelevant phrases, hallucinating protected groups not mentioned in the text, and omitting causal justifications.  Table~\ref{tab:perturbation} shows an example involving the post \textit{``Calling white trash is an insult to styrofoam! The Klan is despicable.''} The original explanation correctly highlights \textit{``white trash''}, links it \textit{White people}, and grounds the label decision causally, resulting in a high \textsf{HateXScore} of 1.0. After perturbation, the metric drops sharply. Substituting the rationale with \textit{``white bubble''} (Perturbed 1) eliminates overlap and causal grounding (QF = CC = 0), yielding a score of 0.5. In Perturbed 2, the explanation retains the slur but fails to identify the target group, reducing TGI and CC, and lowering the score to 0.5. These cases show that \textsf{HateXScore} penalizes explanations that break coherence, misattribute targets, or lose causal clarity, even when the predicted label remains the same.

\section{Conclusion}

We introduced \textsf{HateXScore}, a suite of metrics for evaluating LLM-generated explanations in hate speech detection. Beyond label accuracy, it provides a principled, multifaceted assessment of explanatory quality. Experiments on six datasets and seven models show that \textsf{HateXScore} reveals interpretability failures and annotation inconsistencies overlooked by standard metrics. Notably, when model predictions diverge from labels, higher \textsf{HateXScore} values align with human judgments, underscoring its effectiveness in flagging potentially ambiguous or controversial annotations for further human adjudication.

\section*{Limitations}

Despite its advantages, \textsf{HateXScore} faces certain limitations. It relies on automated methods for identifying and masking rationale text, which can be prone to errors, particularly in highly nuanced or figurative contexts. Moreover, evaluating whether a group is truly attacked remains an inherently subjective judgment, albeit mitigated by structured target-group inventories. In addition, We have been experimenting with English, Chinese and Korean, but more languages are still to be expanded. Finally, \textsf{HateXScore} currently focuses on evaluating static, text-based explanations; future work could explore dynamic or multimodal reasoning (e.g., image-text hate speech) and human–AI co-evaluation frameworks to better capture the evolving nature of online hate.

For instances attacking multiple protected groups, we currently mark TGI as successful if the explanation identifies any clearly targeted group. This simplifies annotation but does not capture target coverage (partial or full), which can overstate performance on multi-target examples. Future work can adopt graded coverage labels or set-based scoring over targeted groups.

% Finally, our experiments have focused on English-language corpora, and future extensions may address multilingual complexities.  

% Since December 2023, a "Limitations" section has been required for all papers submitted to ACL Rolling Review (ARR). This section should be placed at the end of the paper, before the references. The "Limitations" section (along with, optionally, a section for ethical considerations) may be up to one page and will not count toward the final page limit. Note that these files may be used by venues that do not rely on ARR so it is recommended to verify the requirement of a "Limitations" section and other criteria with the venue in question.

\section*{Ethical Statement}

This research focuses on developing and evaluating \textsf{HateXScore} for detecting and explaining hateful or offensive language. Our primary objective is to highlight gaps in current hate-speech detection systems and propose methods (via \textsf{HateXScore}) that improve both classification and explanation quality. We do not seek to facilitate censorship or suppress legitimate free expression. Instead, by revealing vulnerabilities and inconsistencies in detection pipelines, we hope to support more accurate, transparent, and fair content moderation practices.
 
Consistent with the approach in hateful-language research, we rely on established datasets (e.g., HateXplain, HateCheck, Latent Hatred, HASOC, ToxiCN and KOLD). No additional personal or private data was collected for this study. As with other public datasets of toxic or hateful content, all resources are for scientific research only. 
ToxiCN is licensed under a Creative Commons Attribution- NonCommercial-NoDerivatives 4.0 International License (CC BY-NC-ND 4.0). HateCheck and HateXplain are under cc-by-4.0
license.

Examples throughout this paper necessarily include hateful and offensive terms, which may be disturbing or harmful to readers and annotators alike. We have attempted to minimize repeated exposure to such text by limiting the dataset size for examples and advising annotators on content warnings.

Our methods and code are designed to be transparent, allowing researchers, practitioners, and regulatory bodies to understand, audit, and improve our approach. We welcome inquiries, critiques, and collaborations that aim to further the responsible use of these technologies.

By adhering to these principles, we strive to advance the scientific study of hateful language detection and the interpretability of machine learning models in a manner that respects human dignity, privacy, and well-being.

\section*{Acknowledgments}
This research project is supported by the National Research Foundation, Singapore under its National Large Language Models Funding Initiative, (AISG Award No: AISG-NMLP-2024-004), Ministry of Education, Singapore, under its MOE Academic Research Fund Tier 2 (Award No: T2EP20222-0036). Any opinions, findings and conclusions or recommendations expressed in this material are those of the author(s) and do not reflect the views of the National Research Foundation and Ministry of Education, Singapore.

% This document has been adapted
% by Steven Bethard, Ryan Cotterell and Rui Yan
% from the instructions for earlier ACL and NAACL proceedings, including those for
% ACL 2019 by Douwe Kiela and Ivan Vuli\'{c},
% NAACL 2019 by Stephanie Lukin and Alla Roskovskaya,
% ACL 2018 by Shay Cohen, Kevin Gimpel, and Wei Lu,
% NAACL 2018 by Margaret Mitchell and Stephanie Lukin,
% Bib\TeX{} suggestions for (NA)ACL 2017/2018 from Jason Eisner,
% ACL 2017 by Dan Gildea and Min-Yen Kan,
% NAACL 2017 by Margaret Mitchell,
% ACL 2012 by Maggie Li and Michael White,
% ACL 2010 by Jing-Shin Chang and Philipp Koehn,
% ACL 2008 by Johanna D. Moore, Simone Teufel, James Allan, and Sadaoki Furui,
% ACL 2005 by Hwee Tou Ng and Kemal Oflazer,
% ACL 2002 by Eugene Charniak and Dekang Lin,
% and earlier ACL and EACL formats written by several people, including
% John Chen, Henry S. Thompson and Donald Walker.
% Additional elements were taken from the formatting instructions of the \emph{International Joint Conference on Artificial Intelligence} and the \emph{Conference on Computer Vision and Pattern Recognition}.

\bibliography{custom_new}

\appendix

\section{Appendix}
\label{sec:appendix}

\subsection{Dataset Statistics}
\label{dataset}
Table \ref{tab:dataStatistics} shows the statistical distributions of the datasets used in our experiments. Minimal preprocessing was applied (e.g., lowercasing), and each model used its default tokenizer to ensure compatibility. 

\begin{table}[h!]
    \centering
    \small
    % \resizebox{\linewidth}{!}{
    \begin{tabular}{l r r r r }
        \toprule
        \textbf{Dataset} & \textbf{\# Hateful} & \textbf{\#Non-hateful} & \textbf{\# Total} \\
        \midrule
        HateXplain & 688 & 682 & 1,370 \\
        Latent Hatred & 1,092 & 1,772 & 2,864 \\
        HASOC & 807 & 785 & 1,592 \\
        HateCheck & 2,563 & 1,165 & 3,728 \\
        ToxiCN & 1,274 & 1,137 & 2,411 \\
        KOLD & 2,000 & 2,000 & 4,000 \\
        \bottomrule
    \end{tabular}
    % }
\caption{Statistical information for datasets used in our experiments}
\label{tab:dataStatistics}
\end{table}

\begin{table}[ht]
    \centering
    \small
    \resizebox{\linewidth}{!}{
    \begin{tabular}{l}
    \toprule
    \textbf{Text:} \textit{[The contents from datasets]} \\
    \textbf{Reasoning:} \textit{[The analysis from models]} \\
    \midrule
    \textbf{Questions:} \\
    1. What is the score of Quotation Faithfulness (QF)? (0-1) \\
    Answer: \\
    % 2. What is the score of Highlighted Rationales Validation (HRV)? (0-1) \\
    % Answer: \\
    2. What is the score of Target-Group Identification (TGI)? (0-1) \\
    Answer: \\
    3. Decision on Hatefulness (0 $=$ non-hateful, 1 $=$ hateful) \\
    Answer:  \\
        \bottomrule    
    \end{tabular}
    }
\caption{The Instructions Given To Annotators}
\label{tab:human_instruc}
\end{table}

\subsection{Case Study Examples}
\label{cases}
\begin{table*}[htbp]
  \footnotesize
  \centering
  \begin{tabular}{p{1.8cm}llp{8cm}p{2cm}cc}
  % \begin{tabularx}{\textwidth}{@{}J{1.8cm}  M{1.2cm}  M{1cm} J{8cm} M{2cm}@{}}
    \toprule
    \textbf{Text} & \textbf{Label} & \textbf{Pred.} & \textbf{Reasoning} & \textbf{Metrics}\\
    \midrule
    ``It ain't even noon yet and niggas lying.'' &
    Non‑hateful &
    Hateful & 
    To analyze the provided text, we need to consider both the explicit and implicit elements of hate speech. 1. Explicit hate speech: The term \colorbox{yellow}{``niggas''} is a racially charged term that has historically been used as a derogatory term for \colorbox{pink}{black individuals}. Its use can be considered explicit hate speech, as it directly targets a \colorbox{pink}{racial group} with a term that carries a significant amount of historical and social weight. The context in which it is used here suggests a negative connotation, as it is paired with the phrase \colorbox{yellow}{``lying''}, which implies deceitfulness. 2. Implicit hate speech: Implicitly, the use of such a term can perpetuate stereotypes about black individuals, suggesting that they are inherently dishonest or untrustworthy. This reinforces negative societal views and can contribute to a culture of racism and discrimination. The phrase \colorbox{yellow}{``it ain't even noon yet''} could imply that the speaker is surprised or frustrated by the behavior of the individuals they are referring to, which can further imply a judgment based on race. 3. Context and tone: The overall tone of the text appears to express frustration or disbelief, which can be indicative of a broader negative sentiment towards the group being referenced. The casual nature of the language also suggests a normalization of the derogatory term, which can contribute to a culture of hate. In conclusion, the text contains explicit hate speech through the use of a racially derogatory term and implicitly reinforces negative stereotypes about a racial group. Conclusion: The text is hateful. & \makecell[c]{HTC: 1.0\\QF: 1.0\\TGI: 1.0\\CC: 1.0\\HateXScore: 1.0} \\
    \midrule
    ``hate white people.'' &
    Non‑hateful &
    Hateful &
    The text \colorbox{yellow}{``hate white people''} is explicitly expressing a negative sentiment towards \colorbox{pink}{a specific racial group}, which qualifies as hate speech. The phrase directly calls for hatred against individuals based on their race, which is a clear violation of principles promoting respect and equality among different racial and ethnic groups. Upon analyzing the text, we find that it lacks ambiguity or irony. It straightforwardly states a desire to hate a particular group. This explicit declaration of hatred is harmful as it promotes division and animosity based on race, which can lead to further discrimination and social conflict. Given the clear and direct nature of the statement, it does not require interpretation or context to understand its hateful intent. It is a straightforward expression of racial animosity. Conclusion: hateful &
    \makecell[c]{HTC: 1.0\\QF: 1.0\\TGI: 1.0\\CC: 1.0\\HateXScore: 1.0} \\[2pt]
    \midrule
    ``Cut immigration by at least 75\%! Raise does not do enough !'' & Hateful & Non-hateful & Analysis: The text expresses a strong opinion about immigration, advocating for a significant reduction in immigration levels. while it does not contain overtly hateful language or slurs, the statement can imply a negative stance towards \colorbox{pink}{immigrants or immigration as a whole}. This could reflect a belief that immigrants are undesirable or problematic, which can be seen as a form of implicit bias against immigrant communities. However, the text does not explicitly dehumanize or attack any specific group, nor does it use derogatory language.  the phrase \colorbox{yellow}{``raise does not do enough''} is somewhat ambiguous and does not provide clear context about what is being raised or what is insufficient. This could refer to policies or measures related to immigration, but without further context, it is difficult to assess its implications fully. Overall, while the text expresses a strong anti-immigration sentiment, it does not cross the line into hate speech. it reflects a political opinion that may be controversial or divisive, but it does not contain the elements typically associated with hate speech. Conclusion: non-hateful & \makecell[c]{HTC: 1.0\\QF:0.0\\TGI: 1.0\\CC: 0.0\\HateXScore: 0.5} \\
    \bottomrule
  \end{tabular}
  \caption{Several Borderline Examples where the model prediction contradicts the label. Target Group is highlighted by \colorbox{pink}{pink}, Quoted Rationales are highlighted by \colorbox{yellow}{yellow}.}
  
  % Target Group is highlighted by \colorbox{cyan}{cyan}, Quoted Rationales are highlighted by \colorbox{yellow}{yellow}.}
  \label{tab:case_study}
\end{table*}

To illustrate how \textsf{HateXScore} can reveal nuanced explanation behavior beyond simple classification metrics, we focus on several borderline examples from different datasets, as showed in Table~\ref{tab:case_study}, illustrating how classification correctness (HTC) and explanation quality (\textsf{HateXScore}) may sometimes conflict. 
These cases underscore the importance of considering both the label outcome and the underlying rationale, especially in cases with ambiguous or high disagreement. 
\subsubsection{Misclassified Label, Robust Explanation}
The first example, ``it ain't even noon yet and niggas lying.'' is labeled non-hateful in HateXplain. Nonetheless, the model predicts hateful, citing the racial slur and explaining its derogatory connotation. Although the models' prediction disagrees with the dataset annotation, the explanation itself (QF=1, TGI=1) clearly identifies a protected group and shows a significant drop in probability when the slur is masked (QF close to 1). Consequently, the overall \textsf{HateXScore} remains high (0.80). Such a case casts doubt on the ``non-hateful'' label: it is plausible that some annotators considered the usage of ``niggas'' as in-group slang or contextually benign, whereas the model treats it as definitively hateful.
Closer inspection suggested that the annotators themselves might have disagreed on whether the language was truly hateful, underscoring the possibility that the model's explanation exceeded the majority-vote label in clarity or reliability.

% The second example from Latent Hatred states ``: hate white people.'', which is intuitively hateful speech. Yet the dataset's label is non-hateful, likely a misannotation or outlier. The model's explanation quotes ``white people'', identifies race, and exhibits a strong drop in hate probability post-masking (HRV=1). Thus, even though HTC=0, the final \textsf{HateXScore} is relatively high (0.8), signaling a coherent explanation consistent with widely accepted definitions of hateful language. This instance highlights how \textsf{HateXScore} can detect a well-reasoned explanation in the face of questionable labeling.

\subsubsection{Correct Label, Weak Explanation}
In the third example (``cut immigration by at least 75\%! raise does not do enough!''), the label is hateful, presumably reflecting annotators' consensus that targeting ``immigration'' is a discriminatory statement. The model, however, predicts non-hateful. Its explanation references the phrase ``cut immigration,'' but fails to clarify why such a demand may be hateful or to specify any protected group under attack (QF=1, but TGI=0). Despite matching the official label incorrectly, the explanation also scores poorly on reasoning components (e.g., CC = 0), resulting in a lower overall \textsf{HateXScore} (0.50). This example demonstrates that even though the model might show acceptable performance on standard metrics in other samples, it can produce incomplete rationales where content is arguably hateful. \textsf{HateXScore} makes that gap in explanation clarity visible.

These examples highlight how relying solely on Accuracy or F1 can misrepresent a model's behavior, particularly in datasets containing ambiguous, context-dependent, or potentially incorrect labels. While HTC offers the conclusion of prediction, the other \textsf{HateXScore} components (QF, TGI, CC) can give credit to a faithful explanation. Conversely, a model might match the label but fail to justify it in terms of the offending snippet or protected group, leading to a subpar \textsf{HateXScore}.
Overall, these case studies illustrate two core insights: first, classification correctness alone cannot capture explanation completeness and causal relevance; second, labels in hate speech corpora can themselves be flawed or disputed.

\subsection{Instruction for Human Evaluation}
Table~\ref{tab:human_instruc} shows the instructions we gave to the annotator.

\subsection{Sensitivity of the Consistency Threshold ($\tau$)}
\label{sensitive_test}

% Figures~\ref{fig:sensitive_plot}
% show the threshold–sensitivity analysis of \textsf{HateXScore} for six datasets. Each subplot presents the average metric across all models as the threshold $\tau$ varies from 0.1 to 0.9.
% Both metrics decline monotonically as $\tau$ increases, with stable rankings
% around $\tau \approx 0.2$–$0.4$. We thus fix $\tau^* = 0.3$ for all reported
% experiments.

\begin{figure*}[t] 
    \centering
    \begin{tabular}{cc}
    % \begin{tabular}{@{} c@{\hspace{0.01pt}} c@{\hspace{0.05pt}} c @{}}
    \includegraphics[width=0.40\textwidth]{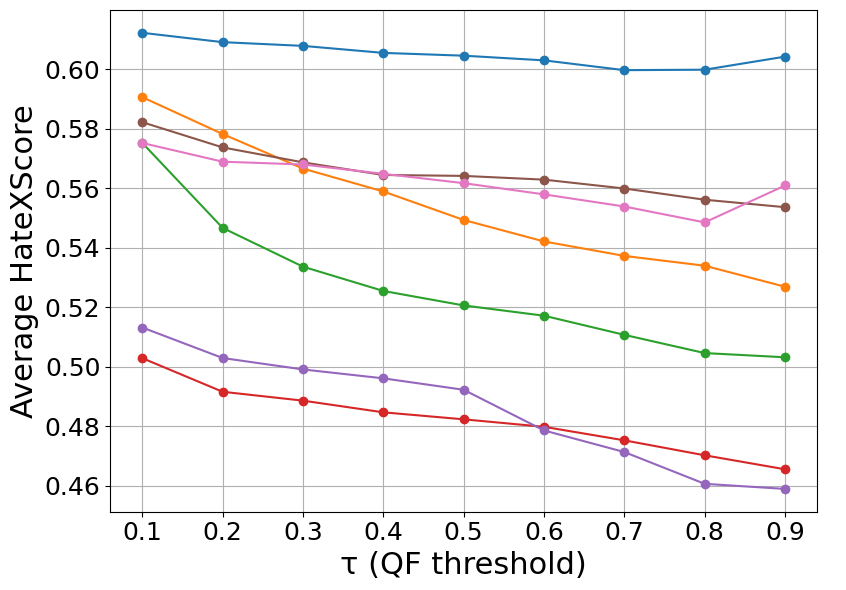} &
    \includegraphics[width=0.55\textwidth]{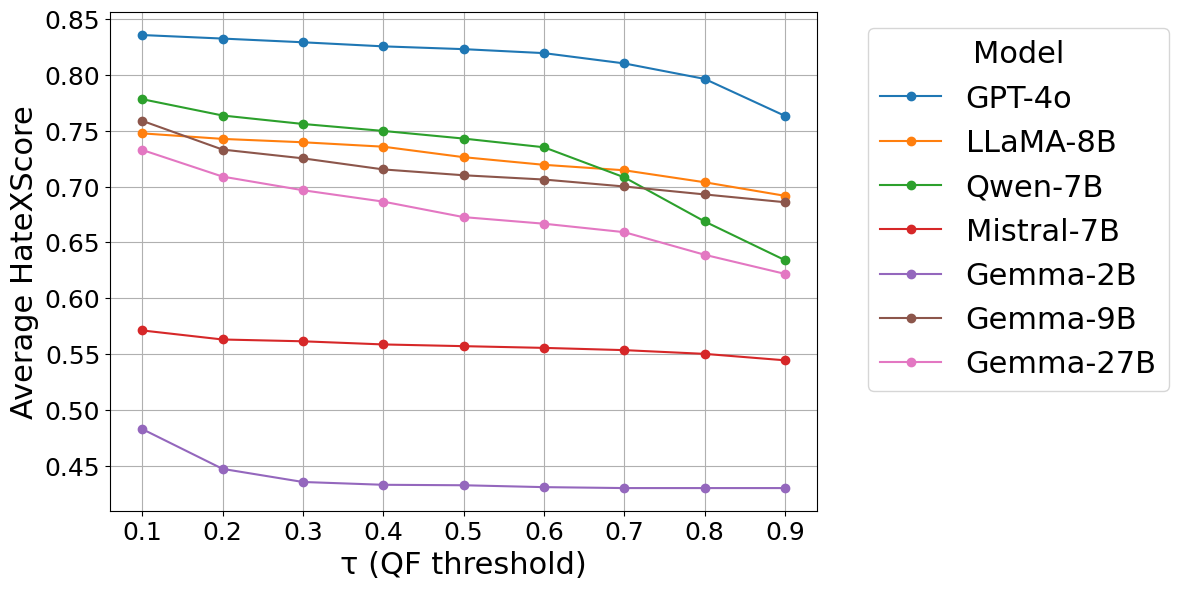} \\
    (a) HASOC & (b) HateCheck \\
\includegraphics[width=0.40\textwidth]{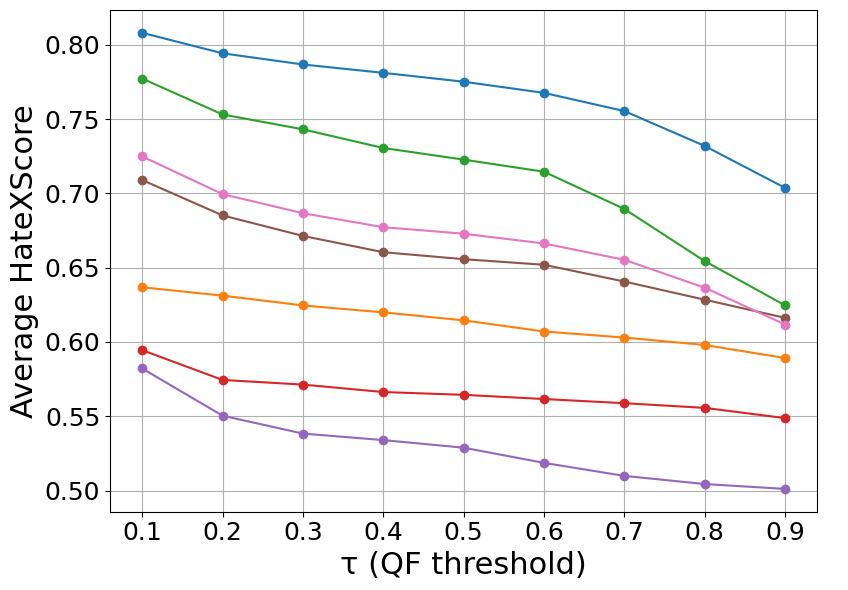} &
\includegraphics[width=0.55\textwidth]{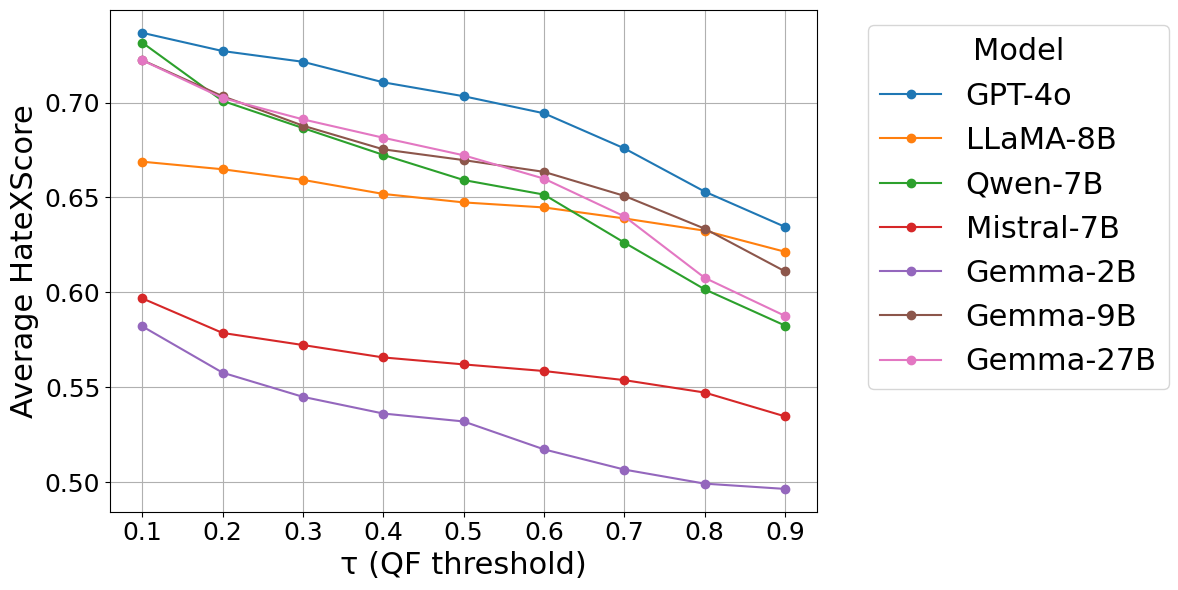} \\
         (c) HateXplain & (d) Latent Hatred \\
    \includegraphics[width=0.40\textwidth]{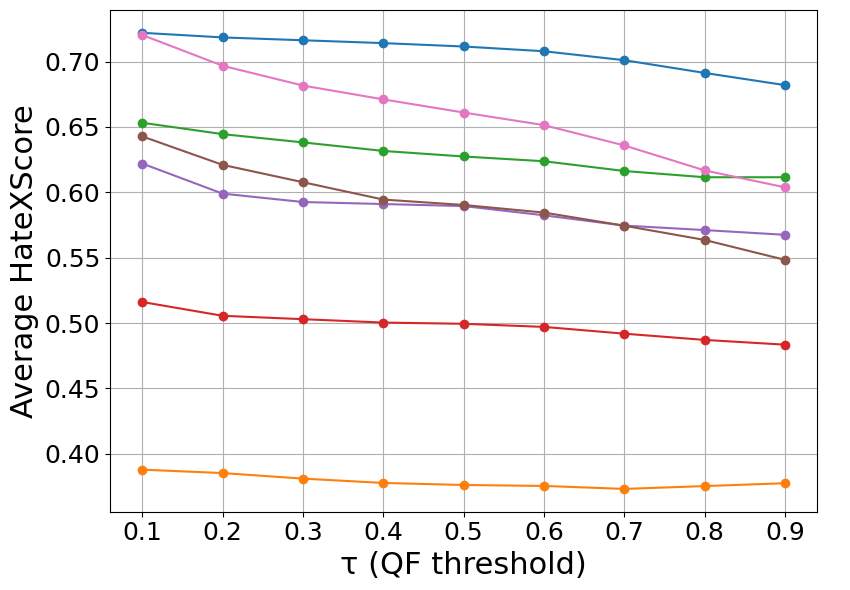} &
    \includegraphics[width=0.55\textwidth]{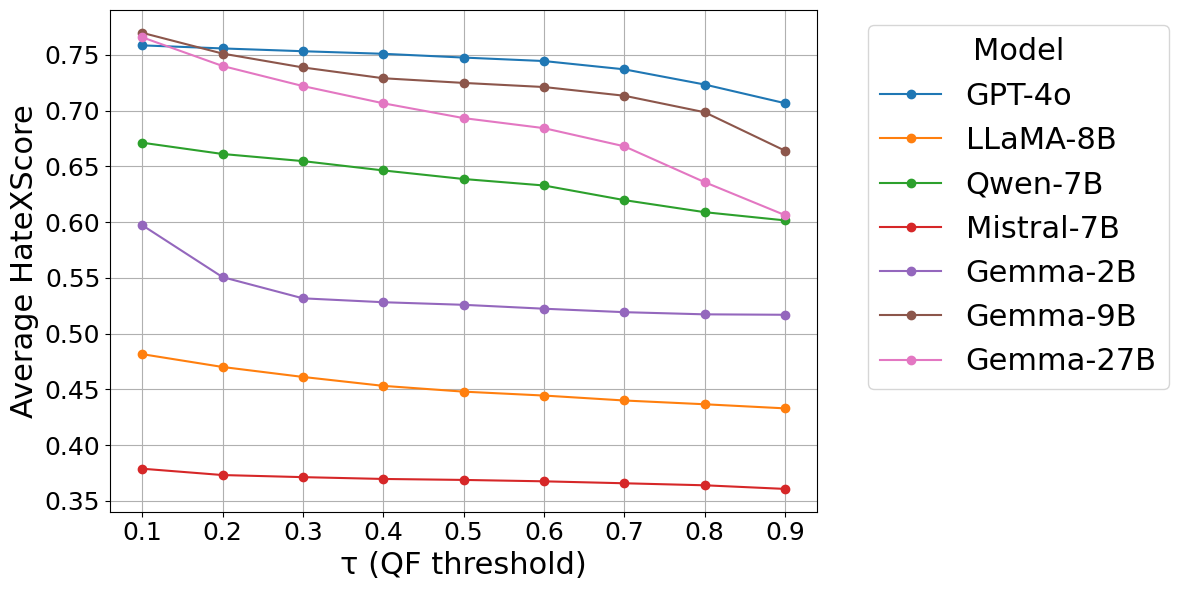} \\
        (e) ToxiCN & (f) KOLD \\
    \end{tabular}
    % \vspace{-10pt}
    \caption{Sensitivity of \textsf{HateXScore} to the Quotation Faithfulness threshold $\tau$ across six datasets. Each subplot shows the average \textsf{HateXScore} for all models as $\tau$ varies from 0.1 to 0.9. Higher curves indicate stronger consistency between explanation and decision.}
    \label{fig:sensitive_plot}
    \vspace{-10pt}
\end{figure*}

We evaluated the robustness of the Consistency Check (CC) and \textsf{HateXScore} metrics by sweeping $\tau$ from 0.1 to 0.9 in steps of 0.1.
Figures~\ref{fig:sensitive_plot} show the average \textsf{HateXScore} for all models across datasets.
As expected, stricter thresholds (larger $\tau$) reduce both CC and \textsf{HateXScore} values, since fewer samples satisfy the consistency condition.
The model ranking, however, remains largely stable for $\tau \leq 0.3$, indicating that the metrics are robust in the low-threshold region.

\begin{table}[h]
\centering
\begin{tabular}{p{0.95\linewidth}}
\toprule
\textbf{\#\#\# Instructions:} \\
% Hate speech is defined as abuse targeted at a protected group or at its members for being a part of that group. \\
Hate speech is defined as any kind of communication in speech, writing, or behaviour that attacks or uses pejorative or discriminatory language with reference to a group or its members on the basis of who they are. \\

Using the above definition, please decide if the following text is hateful or non-hateful. Provide the explanation for your classification in English/Chinese/Korean, followed by the conclusion (Hateful/Non-hateful). In your analysis, consider both explicit and implicit forms of hate speech. Ensure your reasoning avoids over-reliance on keywords and maintains consistency throughout the analysis. \\
\midrule
\textbf{\#\#\# Text:} \\
\{text\} \\
% \textbf{\#\#\# Label:} \\
% \{label\} \\
\bottomrule
\end{tabular}
\caption{}
\label{llm_prompt}
\end{table}

\subsection{Few-shot robustness (2-shot prompting)}
\label{app:2-shot}
Our main experiments focus on the widely used 0-shot deployment setting. Since \textsf{HateXScore} is a metric suite applied to model explanations, it is in principle agnostic to prompting regimes. We therefore conduct a small robustness check to verify that the qualitative trends reported in the main paper remain consistent under light few-shot prompting.

We evaluate the same datasets and models as in Section~\ref{sec:4.2 Models}, but prepend two demonstrations with one hateful and one non-hateful to the prompt used to elicit predictions and explanations. We keep all other settings identical to the main experiments.

Table~\ref{metrics_results_2shot} reports the 2-shot results. Overall, the relative patterns are consistent with the zero-shot setting: models that achieve higher \textsf{HateXScore} under zero-shot prompting also tend to do so under 2-shot prompting, and the separation between label-based performance (Accuracy/F1) and explanation-quality signals (\textsf{HateXScore} and sub-metrics) remains evident across datasets.

\begin{table*}[t!]
\small
\centering
\begin{tabular}{c l c c c c c |c c}
\toprule
\textbf{Dataset} & \textbf{Model} & \textbf{HTC} & \textbf{QF} & \textbf{TGI} & \textbf{CC} & \textbf{HateXScore} & \textbf{Accuracy} & \textbf{F1} \\
\midrule
& GPT-4o & 1.000 & 0.695 & 0.951 & 0.914 & 0.890 & 0.890 & 0.867 \\
& LLaMA-8B & 0.791 & 0.292 & 0.405 & 0.415 & 0.477 & 0.650 & 0.370  \\
& Qwen-7B & 1.000 & 0.620 & 0.926 & 0.841 & 0.846 & 0.758 & 0.737  \\
HateXplain & Mistral-7B & 1.000 & 0.323 & 0.913 & 0.350 & 0.651 & 0.544 & 0.598  \\
& Gemma-2b & 1.000 & 0.288 & 0.820 & 0.454 & 0.640 & 0.647 & 0.675  \\
& Gemma-9b & 1.000 & 0.475 & 0.873 & 0.609 & 0.740 & 0.546 & 0.564  \\
& Gemma-27b & 1.000 & 0.471 & 0.882 & 0.694 & 0.762 & 0.713 & 0.718  \\
\midrule
& GPT-4o & 1.000 & 0.693 & 0.928 & 0.972 & 0.898 & 0.748 & 0.585  \\
& LLaMA-8B & 0.834 & 0.270 & 0.475 & 0.406 & 0.496 & 0.623 & 0.357  \\
& Qwen-7B & 1.000 & 0.643 & 0.905 & 0.887 & 0.858 & 0.694 & 0.605  \\
Latent Hatred & Mistral-7B & 1.000 & 0.392 & 0.848 & 0.493 & 0.685 & 0.638 & 0.646  \\
& Gemma-2b & 1.000 & 0.327 & 0.778 & 0.561 & 0.667 & 0.632 & 0.643  \\
& Gemma-9b & 1.000 & 0.522 & 0.794 & 0.656 & 0.742 & 0.559 & 0.597  \\
& Gemma-27b & 1.000 & 0.556 & 0.802 & 0.793 & 0.788 & 0.694 & 0.668  \\
\midrule
& GPT-4o & 1.000 & 0.885 & 0.851 & 0.996 & 0.933 & 0.504 & 0.057  \\
& LLaMA-8B & 0.810 & 0.387 & 0.425 & 0.565 & 0.547 & 0.503 & 0.139  \\
& Qwen-7B & 1.000 & 0.780 & 0.863 & 0.943 & 0.897 & 0.536 & 0.189  \\
HASOC & Mistral-7B & 0.999 & 0.335 & 0.763 & 0.497 & 0.659 & 0.629 & 0.506  \\
& Gemma-2b & 1.000 & 0.479 & 0.376 & 0.818 & 0.668 & 0.592 & 0.426  \\
& Gemma-9b & 0.997 & 0.537 & 0.475 & 0.705 & 0.679 & 0.565 & 0.428  \\
& Gemma-27b & 1.000 & 0.604 & 0.413 & 0.938 & 0.739 & 0.539 & 0.222  \\
\midrule
& GPT-4o & 1.000 & 0.788 & 0.968 & 0.967 & 0.931 & 0.961 & 0.972  \\
& LLaMA-8B & 0.807 & 0.372 & 0.522 & 0.457 & 0.539 & 0.530 & 0.527  \\
& Qwen-7B & 1.000 & 0.679 & 0.933 & 0.902 & 0.878 & 0.867 & 0.904  \\
HateCheck & Mistral-7B & 1.000 & 0.464 & 0.954 & 0.512 & 0.734 & 0.879 & 0.919  \\
& Gemma-2b & 1.000 & 0.135 & 0.951 & 0.220 & 0.577 & 0.870 & 0.913  \\
& Gemma-9b & 1.000 & 0.756 & 0.892 & 0.814 & 0.865 & 0.873 & 0.913  \\
& Gemma-27b & 1.000 & 0.525 & 0.924 & 0.760 & 0.802 & 0.909 & 0.938  \\
\midrule
& GPT-4o & 1.000 & 0.774 & 0.963 & 0.948 & 0.921 & 0.640 & 0.511  \\
& LLaMA-8B & 0.657 & 0.098 & 0.515 & 0.470 & 0.435 & 0.495 & 0.264  \\
& Qwen-7B & 0.979 & 0.723 & 0.912 & 0.905 & 0.880 & 0.635 & 0.511  \\
ToxiCN & Mistral-7B & 1.000 & 0.326 & 0.345 & 0.498 & 0.546 & 0.746 & 0.762  \\
& Gemma-2b & 0.997 & 0.371 & 0.358 & 0.509 & 0.559 & 0.679 & 0.709  \\
& Gemma-9b & 0.968 & 0.393 & 0.630 & 0.548 & 0.635 & 0.668 & 0.719  \\
& Gemma-27b & 0.987 & 0.487 & 0.910 & 0.766 & 0.787 & 0.695 & 0.701  \\
\midrule
& GPT-4o & 1.000 & 0.772 & 1.000 & 0.977 & 0.937 & 0.677 & 0.529  \\
& LLaMA-8B & 0.502 & 0.274 & 0.758 & 0.471 & 0.501 & 0.505 & 0.231  \\
& Qwen-7B & 0.840 & 0.614 & 0.993 & 0.864 & 0.828 & 0.572 & 0.264  \\
KOLD & Mistral-7B & 1.000 & 0.410 & 0.613 & 0.406 & 0.608 & 0.838 & 0.837  \\
& Gemma-2b & 0.998 & 0.337 & 0.521 & 0.592 & 0.612 & 0.784 & 0.798  \\
& Gemma-9b & 0.977 & 0.608 & 0.849 & 0.758 & 0.798 & 0.788 & 0.790  \\
& Gemma-27b & 0.989 & 0.586 & 0.999 & 0.843 & 0.854 & 0.780 & 0.768  \\
\bottomrule
\end{tabular}
\caption{2-shot prompting robustness check. Macro-averaged HTC/QF/TGI/CC/\textsf{HateXScore} and Accuracy/Macro-F1 across datasets (same evaluation pipeline as Table 1)}
\label{metrics_results_2shot}
\vspace{-10pt}
\end{table*}

% Table C.1: 2-shot prompting robustness check. Macro-averaged HTC/QF/TGI/CC/\textsf{HateXScore} and Accuracy/Macro-F1 across datasets (same evaluation pipeline as Table 1).

\subsection{Target Group List}
\label{tgi_list}
Table~\ref{tgi} - Table~\ref{tgi_zh} shows the default target group lists from the UN~\footnote{https://www.un.org/en/hate-speech/impact-and-prevention/targets-of-hate}, Meta~\footnote{https://transparency.meta.com/en-gb/policies/community-standards/hateful-conduct/}, YouTube~\footnote{https://support.google.com/youtube/answer/2801939?hl=en},  and Twitter~\footnote{https://help.x.com/en/rules-and-policies/hateful-conduct-policy} that we have embedded in our code.

% ~\footnote{https://help.x.com/en/rules-and-policies/hateful-conduct-policy}

\begin{table*}[!ht]
    \centering
    \small
    \begin{tabular}{llp{10cm}ccc}
\toprule
\textbf{Policy} & \textbf{Category} & \textbf{Target Group}\\
\midrule
  & \textbf{National} & afghanistan, albania, algeria, andorra, angola, antigua and barbuda, argentina, armenia, australia, austria, azerbaijan, bahamas, bahrain, bangladesh, barbados, belarus, belgium, belize, benin, bhutan, bolivia, bosnia and herzegovina, botswana, brazil, brunei, bulgaria, burkina faso, burundi, cabo verde, cambodia, cameroon, canada, central african republic, chad, chile, china, colombia, comoros, costa rica, croatia, cuba, cyprus, czech republic, democratic republic of the congo, denmark, djibouti, dominica, dominican republic, ecuador, egypt, el salvador, equatorial guinea, eritrea, estonia, eswatini, ethiopia, federated states of micronesia, fiji, finland, france, gabon, gambia, georgia, germany, ghana, greece, grenada, guatemala, guinea, guinea-bissau, guyana, haiti, honduras, hungary, iceland, india, indonesia, iran, iraq, ireland, israel, italy, jamaica, japan, jordan, kazakhstan, kenya, kiribati, kuwait, kyrgyzstan, laos, latvia, lebanon, lesotho, liberia, libya, liechtenstein, lithuania, luxembourg, madagascar, malawi, malaysia, maldives, mali, malta, marshall islands, mauritania, mauritius, mexico, moldova, monaco, mongolia, montenegro, morocco, mozambique, myanmar, namibia, nauru, nepal, netherlands, new zealand, nicaragua, niger, nigeria, north korea, north macedonia, norway, oman, pakistan, palau, panama, papua new guinea, paraguay, peru, philippines, poland, portugal, qatar, romania, russia, rwanda, saint kitts and nevis, saint lucia, saint vincent and the grenadines, samoa, san marino, sao tome and principe, saudi arabia, senegal, serbia, seychelles, sierra leone, singapore, slovakia, slovenia, solomon islands, somalia, south africa, south korea, south sudan, spain, sri lanka, sudan, suriname, sweden, switzerland, syria, tajikistan, tanzania, thailand, timor-leste, togo, tonga, trinidad and tobago, tunisia, turkey, turkmenistan, tuvalu, uganda, ukraine, united arab emirates, united kingdom, united states of america, uruguay, uzbekistan, vanuatu, vatican city, venezuela, vietnam, yemen, zambia, zimbabwe \\
  \cmidrule{2-3}
   & \textbf{Ethnic} & black, white, asian, latino, jewish, arab, indian, african, nigger, african american, caucasian, hispanic, asian, native american, pacific islander, middle eastern, north african, indigenous, aboriginal \\
  \cmidrule{2-3}
  & \textbf{Religious} & christian, catholic, protestant, orthodox, anglican, baptist, mormon, jehovah's witness, judaism, islam, hinduism, buddhism, sikhism, shinto, taoism, atheist, agnostic, pagan, zoroastrianism, jainism, baha, scientologist, rastafarian, unitarian, falun gong, druze, samaritan, yazidi, ahmadi, alawite, coptic, animist, wiccan, satanist, seventh-day adventist, muslim, jew\\
  \cmidrule{2-3}
\textbf{UN} & \textbf{Disabilities} & disabled, blind, deaf, mute, autistic, down syndrome, schizophrenic, bipolar, mentally ill, wheelchair user, paraplegic, quadriplegic, dwarf, albino, epileptic, diabetic, hiv positive, cancer patient, obese, amputee, retarded \\
  \cmidrule{2-3}
  & \textbf{Sex} & lesbian, gay, bisexual, queer, pansexual, asexual, lgbt \\
  \cmidrule{2-3}
  & \textbf{Migrants} & refugee, immigrant, migrant, asylum seeker, foreigner, expatriate, stateless \\
  \cmidrule{2-3}
  & \textbf{Others} & indigenous peoples, forcibly displaced persons, vocational targets\\
\midrule
 & \textbf{Race or ethnicity} &  black, white, asian, latino, jewish, arab, indian, african, nigger, african american, caucasian, hispanic, asian, native american, pacific islander, middle eastern, north african, indigenous, aboriginal \\
 \cmidrule{2-3}
 & \textbf{National origin} & \textit{[Same as National and Ethnic of UN]} \\
 \cmidrule{2-3}
\textbf{Meta} & \textbf{Disability or serious disease}  & disabled, blind, deaf, mute, autistic, down syndrome, schizophrenic, bipolar, mentally ill, wheelchair user, paraplegic, quadriplegic, dwarf, albino, epileptic, diabetic, hiv positive, cancer patient, obese, amputee, retarded \\
 \cmidrule{2-3}
 & \textbf{Religious affiliation} & christian, catholic, protestant, orthodox, anglican, baptist, mormon, jehovah's witness, judaism, islam, hinduism, buddhism, sikhism, shinto, taoism, atheist, agnostic, pagan, zoroastrianism, jainism, baha, scientologist, rastafarian, unitarian, falun gong, druze, samaritan, yazidi, ahmadi, alawite, coptic, animist, wiccan, satanist, seventh-day adventist, muslim, jew \\
 \cmidrule{2-3}
 &\textbf{ Caste} & dalit, brahmin, kshatriya, vaishya, shudra  \\
\cmidrule{2-3}
 & \textbf{Sexual orientation} & lesbian, gay, bisexual, queer, pansexual, asexual, lgbt \\
 \cmidrule{2-3}
 % & Sex  &  \\
 % \cmidrule{2-3}
 & \textbf{Gender identity} & women, men, transgender, non-binary, intersex, cisgender, female, male \\
 \cmidrule{2-3}
 & \textbf{Immigration} & refugee, immigrant, migrant, asylum seeker, foreigner, expatriate, stateless \\
%  & Race or ethnicity  &  \\
%  \cmidrule{2-3}
%  & National origin  &  \\
%  \cmidrule{2-3}
%  & Religious affiliation  &  \\
%  \cmidrule{2-3}
% Twitter & Sex   &  \\
% \cmidrule{2-3}
%  & Gender identity  &  \\
%  \cmidrule{2-3}
%  & Sexual orientation  &  \\
%  \cmidrule{2-3}
%  & Age    &  children, teenager, youth, adult, senior \\
%  \cmidrule{2-3}
%  & Disability or serious disease  &  \\
% \midrule
%  & Age    &  \\
%  \cmidrule{2-3}
%  & Caste, Ethnicity, or Race    &  \\
%  \cmidrule{2-3}
%  & Disability    &  \\
%  \cmidrule{2-3}
%  & Immigration Status  &  \\
%  \cmidrule{2-3}
% YouTube & Nationality  &  \\
% \cmidrule{2-3}
%  & Religion &  \\
%  \cmidrule{2-3}
%  & Sex, Gender, or Sexual Orientation &  \\
%  \cmidrule{2-3}
%  & Veteran Status &  \\
%  \cmidrule{2-3}
%  & Victims &  \\
\\  \bottomrule
    \end{tabular}
\caption{English Listed target groups of hate speech}
\label{tgi}
    \label{tab:motivate_example}
\end{table*}

\begin{table*}[!ht]
    \centering
    \small
    \begin{tabular}{llp{10cm}ccc}
\toprule
\textbf{Policy} & \textbf{Category} & \textbf{Target Group}\\
\midrule
& \textbf{Race or ethnicity} & black, african american, white, caucasian, hispanic, latino, asian, native american, pacific islander, middle eastern, north african, indigenous, aboriginal \\
\cmidrule{2-3}
& \textbf{National origin} & \textit{[Same as National and Ethnic of UN]} \\
\cmidrule{2-3}
& \textbf{Religious affiliation} & christian, catholic, protestant, baptist, methodist, lutheran, presbyterian, orthodox christian, mormon, jehovah’s witnesses, jewish, muslim, sunni, shia, sufi, ahmadiyya, druze, ismaili, buddhist, hindu, sikh, jain, zoroastrian, baha’i, taoist, confucian, shinto, pagan, atheist, agnostic, humanist \\
\cmidrule{2-3}
\textbf{Twitter} & \textbf{Sex} & male, female, intersex \\
\cmidrule{2-3}
& \textbf{Gender identity} & cisgender man, cisgender woman, transgender man, transgender woman, non-binary, genderqueer, agender, bigender, genderfluid, pangender, two-spirit \\
\cmidrule{2-3}
& \textbf{Sexual orientation} & heterosexual, gay, lesbian, bisexual, pansexual, asexual, demisexual, polysexual, queer \\
\cmidrule{2-3}
& \textbf{Age} & children, teenager, youth, adult, senior \\
\cmidrule{2-3}
& \textbf{Disability or serious disease} & physical disability, visual impairment, hearing impairment, deafness, intellectual disability, autism spectrum disorder, down syndrome, epilepsy, cerebral palsy, muscular dystrophy, multiple sclerosis, parkinson’s disease, alzheimer’s disease, hiv, aids, diabetes, cancer, asthma, heart disease, chronic kidney disease, autoimmune disorders, lupus, crohn’s disease, fibromyalgia, sickle cell disease, hemophilia, thalassemia, celiac disease, dwarfism, clinically obese, arthritis, rheumatoid arthritis, hepatic disease, tuberculosis, bipolar disorder, depression, anxiety disorders, borderline personality disorder, schizophrenia, aphasia, dyspraxia \\
\midrule
& \textbf{Age} & children, teenager, youth, adult, senior \\
\cmidrule{2-3}
& \textbf{Caste, Ethnicity, or Race} & dalit, adivasi, brahmin, kshatriya, vaishya, shudra, black or african american, white or caucasian, hispanic or latino, asian, native american, pacific islander, middle eastern, north african, indigenous, aboriginal \\
\cmidrule{2-3}
& \textbf{Disability} & physical disability, visual impairment, hearing impairment, deafness, intellectual disability, autism spectrum disorder, epilepsy, cerebral palsy, muscular dystrophy, multiple sclerosis, parkinson’s disease, alzheimer’s disease, hiv, aids, diabetes, cancer, asthma, heart disease, chronic kidney disease, autoimmune disorders, lupus, crohn’s disease, fibromyalgia, sickle cell disease, hemophilia, thalassemia, celiac disease, dwarfism, clinically obese, arthritis, rheumatoid arthritis, hepatic disease, tuberculosis, bipolar disorder, depression, anxiety disorders, borderline personality disorder, schizophrenia, aphasia, dyspraxia \\
\cmidrule{2-3}
& \textbf{Immigration Status} & citizen, permanent resident, documented immigrant, refugee, asylum seeker, undocumented immigrant \\
\cmidrule{2-3}
\textbf{YouTube} & \textbf{Nationality} & \textit{[Same as National and Ethnic of UN]} \\
\cmidrule{2-3}
& \textbf{Religion} & christian, catholic, protestant, baptist, methodist, lutheran, presbyterian, orthodox christian, mormon, jehovah’s witnesses, jewish, muslim, sunni, shia, sufi, ahmadiyya, druze, ismaili, buddhist, hindu, sikh, jain, zoroastrian, baha’i, taoist, confucian, shinto, pagan, atheist, agnostic, humanist \\
\cmidrule{2-3}
& \textbf{Sex, Gender, or Sexual Orientation} & male, female, intersex, cisgender man, cisgender woman, transgender man, transgender woman, non-binary, genderqueer, agender, bigender, genderfluid, pangender, two-spirit, heterosexual, gay, lesbian, bisexual, pansexual, asexual, demisexual, polysexual, queer \\
\cmidrule{2-3}
& \textbf{Veteran Status} & military veteran, non-veteran \\
\cmidrule{2-3}
& \textbf{Victims} & victim of sexual assault, victim of domestic violence, victim of crime, victim of harassment, victim of trafficking, victim of bullying, victim of discrimination 
\\  \bottomrule
    \end{tabular}
    \caption{English Listed target groups of hate speech}
\end{table*}

\begin{table*}[!ht]
    \centering
    \small
    \begin{tabular}{llp{11cm}}
\toprule
 & \textbf{Category} & \textbf{Target Group}\\
\midrule
  & \begin{CJK}{UTF8}{mj} 여성과 소녀 \end{CJK} &  \begin{CJK}{UTF8}{mj} 여성, 여자, 소녀, 여아, 여학생, 여성 인권옹호자, 여성인권옹호자, 여성 정치인, 여성 언론인, 여성 활동가 \end{CJK} \\
  \cmidrule{2-3}
  & \begin{CJK}{UTF8}{mj} 종교적 소수자 \end{CJK} & \begin{CJK}{UTF8}{mj} 종교 소수자, 무슬림, 이슬람교도, 유대인, 유대교도, 시크교도, 힌두교도, 불교도, 바하이 신도, 야지디, 아흐마디야 신도, 기독교 소수파, 소수파 기독교인, 소수 종파 신도 \end{CJK} \\
  \cmidrule{2-3}
  & \begin{CJK}{UTF8}{mj} 인종 민족 국가적 소수자 \end{CJK} &  \begin{CJK}{UTF8}{mj} 인종 소수자, 민족 소수자, 국가적 소수자, 국적 소수자, 흑인, 아프리카계, 라틴계, 라티노, 라티나, 라틴엑스, 아시아계, 동남아계, 중국계, 한국계, 일본계, 중동 북아프리카계, 아랍인, 쿠르드인, 로마인, 롬인, 팔레스타인인, 로힝야, 유럽 소수 민족, 유럽의 소수 민족, 원주민, 선주민, 토착민, 아메리카 원주민, 아보리지니, 토레스 해협 섬 주민, 마오리, 사미, 아이누, 인디헤나, 라틴아메리카 원주민, 라틴 아메리카 원주민, 아프가니스탄, 알바니아, 알제리, 안도라, 앙골라, 안티구아 바부다, 아르헨티나, 아르메니아, 오스트레일리아(호주), 오스트리아, 아제르바이잔, 바하마, 바레인, 방글라데시, 바베이도스, 벨라루스, 벨기에, 벨리즈, 베냉, 부탄, 볼리비아, 보스니아 헤르체고비나, 보츠와나, 브라질, 브루나이, 불가리아, 부르키나파소, 부룬디, 카보베르데, 캄보디아, 카메룬, 캐나다, 중앙아프리카공화국, 차드, 칠레, 중국, 콜롬비아, 코모로, 코스타리카, 크로아티아, 쿠바, 키프로스, 체코(체코공화국), 콩고민주공화국, 덴마크, 지부티, 도미니카, 도미니카공화국, 에콰도르, 이집트, 엘살바도르, 적도기니, 에리트레아, 에스토니아, 에스와티니, 에티오피아, 미크로네시아연방, 피지, 핀란드, 프랑스, 가봉, 감비아, 조지아, 독일, 가나, 그리스, 그레나다, 과테말라, 기니, 기니비사우, 가이아나, 아이티, 온두라스, 헝가리, 아이슬란드, 인도, 인도네시아, 이란, 이라크, 아일랜드, 이스라엘, 이탈리아, 자메이카, 일본, 요르단, 카자흐스탄, 케냐, 키리바시, 쿠웨이트, 키르기스스탄, 라오스, 라트비아, 레바논, 레소토, 라이베리아, 리비아, 리히텐슈타인, 리투아니아, 룩셈부르크, 마다가스카르, 말라위, 말레이시아, 몰디브, 말리, 몰타, 마셜제도, 모리타니, 모리셔스, 멕시코, 몰도바, 모나코, 몽골, 몬테네그로, 모로코, 모잠비크, 미얀마, 나미비아, 나우루, 네팔, 네덜란드, 뉴질랜드, 니카라과, 니제르, 나이지리아, 북한, 북마케도니아, 노르웨이, 오만, 파키스탄, 팔라우, 파나마, 파푸아뉴기니, 파라과이, 페루, 필리핀, 폴란드, 포르투갈, 카타르, 루마니아, 러시아, 르완다, 세인트키츠 네비스, 세인트루시아, 세인트빈센트 그레나딘, 사모아, 산마리노, 상투메 프린시페, 사우디아라비아, 세네갈, 세르비아, 세이셸, 시에라리온, 싱가포르, 슬로바키아, 슬로베니아, 솔로몬제도, 소말리아, 남아프리카공화국, 대한민국(남한), 남수단, 스페인, 스리랑카, 수단, 수리남, 스웨덴, 스위스, 시리아, 타지키스탄, 탄자니아, 태국, 동티모르(티모르레스트), 토고, 통가, 트리니다드 토바고, 튀니지, 튀르키예(터키), 투르크메니스탄, 투발루, 우간다, 우크라이나, 아랍에미리트, 영국, 미국, 우루과이, 우즈베키스탄, 바누아투, 바티칸시국, 베네수엘라, 베트남, 예멘, 잠비아, 짐바브웨 \end{CJK}
 \\
  \cmidrule{2-3}
  & \begin{CJK}{UTF8}{mj} 언어적 소수자 \end{CJK} &  \begin{CJK}{UTF8}{mj} 언어 소수자, 소수 언어 사용자, 이중언어 화자, 사미어 사용자, 아이누어 사용자, 쿠르드어 사용자, 베르베르어 사용자, 아마지그어 사용자, 카탈루냐어 사용자, 카탈란어 사용자, 한국수어 사용자, 수어 사용자  \end{CJK} \\
  \cmidrule{2-3}
  & \begin{CJK}{UTF8}{mj} 이주민 난민 무국적자 \end{CJK} &  \begin{CJK}{UTF8}{mj} 이주민, 이민자, 이주 노동자, 이주노동자, 난민, 난민 신청자, 망명 신청자, 망명신청자, 국내 실향민, 국내실향민, IDP, 무국적자, 미등록 이주민, 미등록이주민, 이주 배경 청년, 이주배경 청년 \end{CJK} \\
  \cmidrule{2-3}
  &  LGBTIQ+ &  \begin{CJK}{UTF8}{mj} 성소수자, LGBTIQ+, 레즈비언, 게이, 양성애자, 바이섹슈얼, 팬섹슈얼, 범성애자, 무성애자, 에이섹슈얼, 트랜스젠더, 트랜스 여성, 트랜스여성, 트랜스 남성, 트랜스남성, 논바이너리, 비이분법, 젠더 비순응, 젠더비순응, 인터섹스, 퀴어 \end{CJK} \\
  \cmidrule{2-3}
  & \begin{CJK}{UTF8}{mj} 장애인 \end{CJK} &  \begin{CJK}{UTF8}{mj} 장애인, 지체장애인, 뇌병변장애인, 시각장애인, 청각장애인, 농인, 난청인, 언어장애인, 지적장애인, 발달장애인, 자폐 스펙트럼 당사자, 자폐스펙트럼 당사자, 자폐성 장애인, 학습장애 당사자, 난독증 당사자, 정신장애인, 정신 건강 장애가 있는 사람, 정신건강 장애가 있는 사람, 희귀질환 장애인, 만성질환 장애인 \end{CJK} \\
  \cmidrule{2-3}
  & \begin{CJK}{UTF8}{mj} 언론인과 인권옹호자 \end{CJK} &  \begin{CJK}{UTF8}{mj} 언론인, 기자, 보도진, 편집자, 팩트체커, 여성 언론인, 여성언론인, 인권옹호자, 인권 활동가, 인권활동가, 시민사회 활동가, 시민사회활동가, 법률 지원 활동가, 환경 운동가, 환경운동가, LGBTIQ+ 인권 활동가, LGBTIQ+ 인권활동가, 여성 인권 활동가, 여성 인권활동가, 여성인권 활동가, 여성인권활동가 \end{CJK} \\
 \bottomrule
    \end{tabular}
\caption{Korean Listed target groups of hate speech}
\label{tgi_kr}
\end{table*}

\begin{table*}[ht]
    \centering
    \small
    \begin{tabular}{llp{13cm}}
\toprule
 & \textbf{Category} & \textbf{Target Group}\\
\midrule
  & 种族 & 非洲人, 非裔美国人, 加勒比非洲裔, 阿拉伯人, 亚美尼亚人, 亚洲人, 亚述人,澳大利亚土著, 巴尔干人, 孟加拉人, 巴斯克人, 柏柏尔人, 黑人, 巴西人,英国人, 保加利亚人, 缅甸人, 高加索人, 柬埔寨人, 卡津人, 中国人, 古巴人,捷克人, 丹麦人, 多米尼加人, 荷兰人, 埃及人, 英格兰人, 爱沙尼亚人, 菲律宾人,芬兰人, 法国人, 格鲁吉亚人, 德国人, 希腊人, 吉普赛人/罗姆人, 海地人, 汉族,西班牙裔, 匈牙利人, 冰岛人, 印度人, 土著, 因纽特人, 伊朗人, 伊拉克人,爱尔兰人, 以色列人, 意大利人, 牙买加人, 日本人, 犹太人, 约旦人, 韩国人,库尔德人, 老挝人, 拉丁裔, 黎巴嫩人, 马来人, 毛利人, 墨西哥人, 中东人,蒙古人, 摩洛哥人, 穆斯林, 美洲原住民, 新西兰人, 尼日利亚人, 北欧人,挪威人, 巴基斯坦人, 巴勒斯坦人, 波斯人, 波兰人, 葡萄牙人, 波多黎各人,罗马尼亚人, 俄罗斯人, 沙特阿拉伯人, 苏格兰人, 塞尔维亚人, 新加坡人, 索马里人,南非人, 西班牙人, 斯里兰卡人, 苏丹人, 瑞典人, 瑞士人, 叙利亚人, 台湾人,泰国人, 藏族人, 土耳其人, 乌克兰人, 越南人, 威尔士人, 也门人, 波斯尼亚人, 白人, 拉丁裔, 非洲人, 黑鬼, 白俄罗斯人, 哈萨克人, 乌兹别克人\\
  \cmidrule{2-3}
   & 宗教 & 基督徒, 天主教徒, 新教徒, 东正教徒, 圣公会, 浸礼会, 摩门教徒,耶和华见证人, 犹太教, 伊斯兰教, 印度教, 佛教, 锡克教, 神道教,道教, 无神论者, 不可知论者, 异教徒, 祆教, 耆那教, 巴哈伊教,科学教, 拉斯塔法里教徒, 一神论者, 法轮功, 德鲁兹教徒, 撒马利亚人,雅兹迪教徒, 艾哈迈迪派, 阿拉维派, 科普特教徒, 万物有灵论者, 威卡教徒, 撒旦教徒,基督复临安息日会 \\
  \cmidrule{2-3}
  & 性别 & 女性, 男性, 跨性别, 非二元, 双性人, 顺性别, 女, 男\\
  \cmidrule{2-3}
 & 性取向 & 女同性恋, 男同性恋, 双性恋, 酷儿, 泛性恋, 无性恋, LGBT \\
  \cmidrule{2-3}
  & 残疾 & 残障人士, 盲人, 聋人, 哑巴, 自闭症患者, 唐氏综合症患者, 精神分裂症患者,躁郁症患者, 精神病患者, 轮椅使用者, 截瘫患者, 四肢瘫痪者, 侏儒,白化病患者, 癫痫患者, 糖尿病患者, 艾滋病毒携带者, 癌症患者, 肥胖者, 截肢者 \\
  \cmidrule{2-3}
  & 年龄 & 儿童, 青少年, 青年, 成年人, 老年人 \\
  \cmidrule{2-3}
  & 移民 & 难民, 移民, 迁徙者, 寻求庇护者, 外国人, 侨民, 无国籍者\\
 \bottomrule
    \end{tabular}
\caption{Chinese Listed target groups of hate speech}
\label{tgi_zh}
\end{table*}

\subsection{LLMs Prompt Formulations}
\label{app:llm_prompt}
Table~\ref{llm_prompt} details the prompt templates used for generating model outputs across all experiments. Each prompt includes a standardized definition of hate speech, followed by an instruction for the model to classify the text as hateful or non-hateful while providing a justification. 
% The section ensures reproducibility of the experimental setup and clarifies the linguistic structure used to elicit consistent reasoning behavior across different large language models.

\end{CJK*}
\end{document}